\documentclass[journal]{IEEEtran}

\usepackage{subcaption}
\usepackage{graphicx}
\usepackage{color,soul}
\usepackage{gensymb}
\usepackage{amssymb}
\usepackage{amsmath} 

\usepackage{tabu}
\usepackage{float}        % For using the [H] placement option
\usepackage{algorithm}
\usepackage{algpseudocode}
\usepackage{makecell}
\usepackage{multicol}
\usepackage{multirow}
\usepackage{enumitem}
\usepackage{xcolor}
\usepackage{url}

\usepackage{xstring}
\usepackage{url}

\newcommand{\revonebibkeys}{,Peng2019,Yu2020,Hu2019b,Li2019CDPN:Estimation,Park2019a,Zakharov2019,Xiang,Beyer2015,Bukschat2020,Wang2021d,Di2021,Liu2025a,Sundermeyer2023,}
\newcommand{\revtwobibkeys}{,Zhang2025a,Yu2024a,Longhini2025,Zhu2025,Zhou2025,DeGusseme2024BenchmarkingGrasp,Caruana1997,Szegedy2014,}
\newcommand{\revthreebibkeys}{,huang_task_2023,chen_deformpam_2025,ha_flingbot_2021,sunil_visuotactile_2022,sunil_reactive_2025,rome_learning_2026,wu2025economic,generalizinggrasp2024,negativepromptguidance2024,regionnormalizedgrasp2024,zerograsp2025,graspcot2025,}
\newcommand{\revfivebibkeys}{,Xue2023,Caporali2020,Zhu2023,Qian2020,Gabas2021,Avigal2022,Lips2022,Wang2021g,Yu2020,Hu2019b,Park2019a,Zakharov2019,Bukschat2020,Wang2021d,Di2021,Sundermeyer2023,Wang2023e,}
\makeatletter
\let\orig@bibitem\bibitem
\let\orig@thebibliography\thebibliography
\let\orig@endthebibliography\endthebibliography
\def\rev@bibendkey{REVFIVEBIBEND}
\long\def\rev@bibstart#1\bibitem#2{#1\rev@bibitem{#2}}
\long\def\rev@bibitem#1#2\bibitem#3{%
    \IfSubStr{\revfivebibkeys}{,#1,}{%
        \rev@removedbibitem{#1}{#2}%
    }{%
    \orig@bibitem{#1}%
    \IfSubStr{\revonebibkeys}{,#1,}{{\color{revonecolor}#2}}{%
    \IfSubStr{\revtwobibkeys}{,#1,}{{\color{revtwocolor}#2}}{%
    \IfSubStr{\revthreebibkeys}{,#1,}{{\color{revthreecolor}#2}}{#2}%
    }}}%
    \def\rev@nextkey{#3}%
    \ifx\rev@nextkey\rev@bibendkey
    \else
        \expandafter\rev@bibitem\expandafter{#3}%
    \fi
}
\RenewDocumentEnvironment{thebibliography}{m +b}{%
    \orig@thebibliography{#1}%
    \expandafter\rev@bibstart#2\bibitem{REVFIVEBIBEND}%
}{%
    \orig@endthebibliography
}

\newif\ifshowrevonechanges
\newif\ifshowrevtwochanges
\newif\ifshowrevthreechanges
\newif\ifshowrevfourchanges
\newif\ifshowrevfivechanges
\showrevonechangesfalse    % hide revision 1 changes in main manuscript
\showrevtwochangesfalse    % hide revision 2 changes in main manuscript
\showrevthreechangesfalse  % hide revision 3 changes in main manuscript
\showrevfourchangesfalse   % hide revision 4 changes in main manuscript
\showrevfivechangesfalse   % hide revision 5 changes in main manuscript

\colorlet{revonebase}{magenta!75!black}
\colorlet{revtwobase}{teal!70!black}
\colorlet{revthreebase}{blue}
\colorlet{revfourbase}{blue}
\colorlet{revfivebase}{blue}
\ifshowrevonechanges
  \colorlet{revonecolor}{revonebase}
\else
  \colorlet{revonecolor}{black}
\fi
\ifshowrevtwochanges
  \colorlet{revtwocolor}{revtwobase}
\else
  \colorlet{revtwocolor}{black}
\fi
\ifshowrevthreechanges
  \colorlet{revthreecolor}{revthreebase}
\else
  \colorlet{revthreecolor}{black}
\fi
\ifshowrevfourchanges
  \colorlet{revfourcolor}{revfourbase}
\else
  \colorlet{revfourcolor}{black}
\fi
\ifshowrevfivechanges
  \colorlet{revfivecolor}{revfivebase}
\else
  \colorlet{revfivecolor}{black}
\fi

\newcommand{\revtwochanges}[1]{{\color{revtwocolor}#1}}
\newcommand{\revthreechanges}[1]{{\color{revthreecolor}#1}}
\newcommand{\revfourchanges}[1]{{\color{revfourcolor}#1}}
\newcommand{\revfivechanges}[1]{{\color{revfivecolor}#1}}
\ExplSyntaxOn
\seq_new:N \l__rev_strike_words_seq
\cs_new_protected:cpn { rev@strikewords } #1
  {
    \seq_set_split:Nnn \l__rev_strike_words_seq { ~ } {#1}
    \seq_map_indexed_inline:Nn \l__rev_strike_words_seq
      {
        \int_compare:nNnT {##1} > {1} {\space}
        \sout{##2}
      }
  }
\ExplSyntaxOff
\long\def\rev@strikevenue#1\emph#2#3\rev@endvenue{%
  \rev@strikewords{#1}\space
  {\itshape\rev@strikewords{#2}}%
  \rev@strikewords{#3}%
}
\long\def\rev@strikeentry#1``#2,''#3\par\rev@endentry{%
  \rev@strikewords{#1}\space
  \sout{``}\rev@strikewords{#2}\sout{,''}\space
  \rev@strikevenue#3\rev@endvenue
}
\long\def\rev@strikeentrynoemph#1``#2,''#3\par\rev@endentry{%
  \rev@strikewords{#1}\space
  \sout{``}\rev@strikewords{#2}\sout{,''}\space
  \rev@strikewords{#3}%
}
\ifshowrevfivechanges
  \newcommand{\rev@removedbibitem}[2]{%
    \begingroup
      \color{revfivecolor}%
      \let\rev@origbiblabel\@biblabel
      \def\@biblabel##1{\sout{\rev@origbiblabel{##1}}}%
      \orig@bibitem{#1}%
      \IfStrEq{#1}{Bukschat2020}{%
        \rev@strikeentrynoemph#2\rev@endentry
      }{%
        \rev@strikeentry#2\rev@endentry
      }%
    \endgroup
  }
  \newcommand{\revfiveremovedcite}[1]{{\color{revfivecolor}\sout{\cite{#1}}}\hspace{0.3em}}
\else
  \newcommand{\rev@removedbibitem}[2]{\orig@bibitem{#1}#2}
  \newcommand{\revfiveremovedcite}[1]{}
\fi

\makeatother

\usepackage{booktabs}

\title{Vision-Based 6-DoF Grasp Pose Estimation for Robot Cloth Unfolding}

\author{
    \IEEEauthorblockN{
    Domen Tabernik\IEEEauthorrefmark{2}\textsuperscript{*},\ 
    Peter Nimac\IEEEauthorrefmark{3}\IEEEauthorrefmark{4}\textsuperscript{*},\ 
    Jan Jeri\'cevi\'c\IEEEauthorrefmark{3}\IEEEauthorrefmark{4},\ 
    Danijel Sko\v{c}aj\IEEEauthorrefmark{2},\space
    Andrej Gams\IEEEauthorrefmark{4}
  }
  
  \IEEEauthorblockA{%
    \IEEEauthorrefmark{2}%
      Faculty of Computer and Information Science, University of Ljubljana, Ve\v{c}na pot 113, 1000 Ljubljana, Slovenia\\
      \{domen.tabernik, danijel.skocaj\}@fri.uni-lj.si
  }%   
  
  \IEEEauthorblockA{%
    \IEEEauthorrefmark{3}%
      Jo\v{z}ef Stefan International Postgraduate School, Jamova cesta 39, 1000 Ljubljana, Slovenia \\
  }%
  
  \IEEEauthorblockA{%
    \IEEEauthorrefmark{4}%
      Jo\v{z}ef Stefan Institute, Jamova cesta 39, 1000 Ljubljana, Slovenia \\
      \{peter.nimac, jan.jericevic, andrej.gams\}@ijs.si
  }%
 \thanks{\textsuperscript{*}Domen Tabernik and Peter Nimac contributed equally to this work as first authors.}
}

\begin{document}

\maketitle

\begin{abstract}
Cloth manipulation is a challenging task due to the deformable and high-dimensional nature of cloth, which leads to complex interaction dynamics and perceptual ambiguity arising from frequent occlusions of critical visual cues such as folds, edges, and grasp points. In this work, we tackle cloth unfolding using a regrasping-in-the-air strategy, where one manipulator holds the cloth while the other grasps it at an optimally selected point to unfold it. To this end, we propose CeDiRNet-6DoF, a deep learning framework that jointly predicts effective grasp points and the complete 6-DoF grasp pose from the observed cloth configuration. By integrating dense 3D grasp regression with segmentation and sine–cosine-encoded Euler angles, the proposed method reliably estimates the grasp configuration that maximi\revfivechanges{z}es the unfolded cloth area. We extensively evaluated CeDiRNet-6DoF on a bimanual robotic setup within the ICRA 2024 Cloth Competition framework, achieving state-of-the-art performance. An ablation study further validates the benefits of key design components, including joint segmentation, background randomization, and image cropping. These results establish CeDiRNet-6DoF as a robust and versatile foundation for reliable robotic cloth manipulation in unstructured environments.

\end{abstract}

\begin{IEEEkeywords}
Robotic Manipulation, Cloth Manipulation, Grasp Pose Regression, 6-DoF, Deep Learning.
\end{IEEEkeywords}

\section{Introduction}
Robotic manipulation of deformable objects, such as cloth and textiles, represents a fundamental yet challenging problem in modern robotics. Unlike rigid objects\revtwochanges{, for which robotic grasping~\cite{Zhang2025a} or pushing~\cite{Yu2024a} typically results in a well-predicted object state,} cloths exhibit high degrees of freedom, complex non-linear dynamics, and significant variability in shape and appearance\revtwochanges{~\cite{Longhini2025}}. These challenges make tasks like folding, unfolding, and precise grasping particularly difficult to automate, especially in unstructured environments\revtwochanges{~\cite{Zhu2025}}. %Overcoming these hurdles is essential for a wide range of applications including automated laundry systems, garment manufacturing, and even delicate surgical procedures involving soft tissues.

Folding and unfolding are fundamental operations in cloth manipulation. Prior work has explored a range of strategies—from bimanual techniques~\revfiveremovedcite{Xue2023}\cite{Barbany2025} and flinging\revtwochanges{~\cite{Zhou2025}} to regrasping-in-the-air~\cite{Maitin-Shepard2010,Doumanoglou2014,DeGusseme2024BenchmarkingGrasp}. In this paper, we focus on unfolding as a critical precursor to effective folding. Specifically, we adopt a regrasping-in-the-air strategy where one robot arm suspends the cloth while the other selects the optimal grasp point to achieve full unfolding. The ICRA 2024 Cloth Competition\revtwochanges{~\cite{DeGusseme2024BenchmarkingGrasp}} recently introduced a standardized dual-arm benchmark for this task, enabling real-world evaluation. By following this framework, our approach facilitates robust and reproducible comparisons with state-of-the-art methods.

%Cloth manipulation stands as one of the most challenging tasks in robotics, with folding and unfolding playing a central role in a wide array of applications. Various strategies have been proposed—from coordinated bimanual methods and dynamic flinging to more sophisticated airborne regrasping techniques. In our work, we focus on the unfolding process, a critical prerequisite not only for accurate folding but also for executing effective regrasping in the air. The ICRA 2024 Cloth Competition has introduced a rigorously defined task, complete with standardized setups and evaluation protocols, addressing the longstanding difficulties in replicating real-world conditions and benchmarking across different methods. While simulation environments such as GarmentLab offer some insights, they often fail to capture the subtle nuances of fabric—like rigidity and seam details—that can significantly impact grasping performance. By adhering to the ICRA 2024 framework, our approach leverages the regrasping-in-the-air strategy to enable the most robust and standardized evaluation possible, thereby facilitating direct and meaningful comparisons with state-of-the-art techniques in real-world setups.

\begin{figure}
    \centering
    \includegraphics[width=1\linewidth]{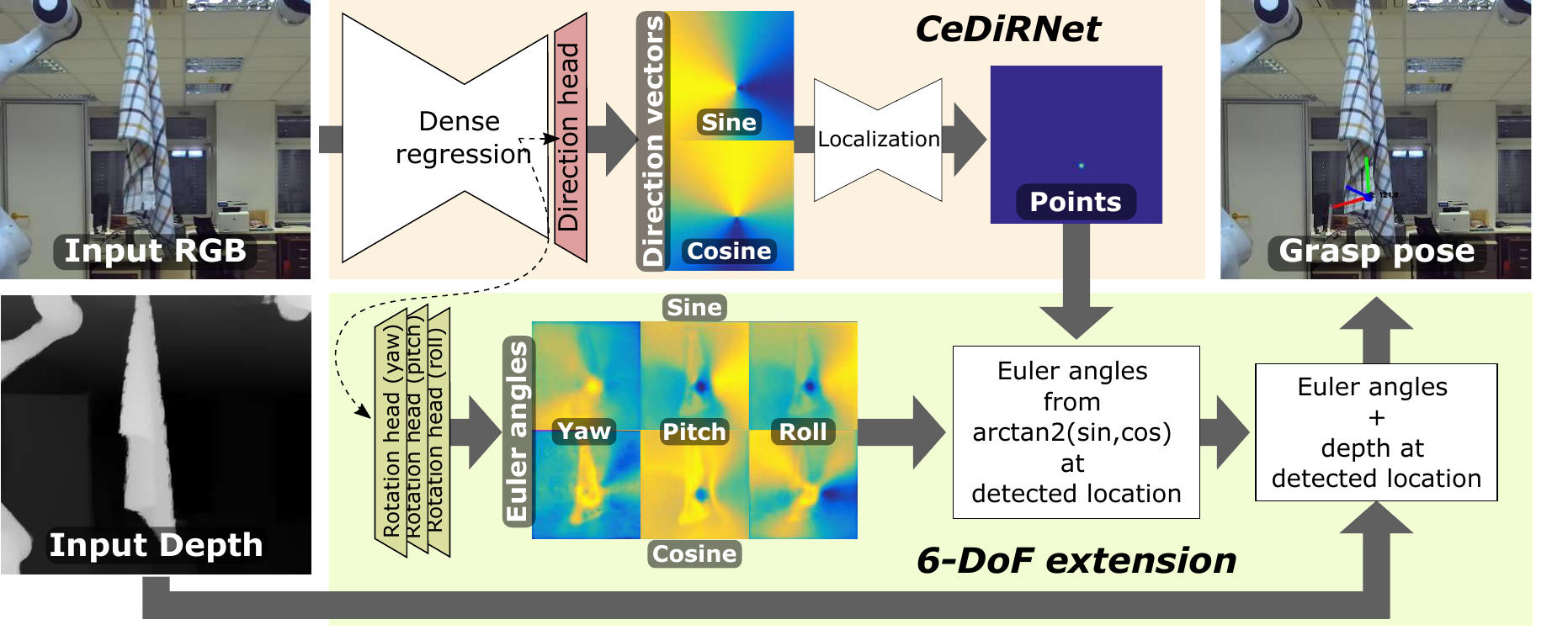}
    \caption{\revfourchanges{CeDiRNet-6DoF architecture. CeDiRNet (top) regresses center directions for 2D grasp localization; its 6-DoF extension (bottom) \revthreechanges{regresses sine/cosine pairs for roll, pitch, and yaw.} Orientation and depth yield the 6-DoF pose.}}
    \label{fig:CeDiRNet-6DoF-arch}
\end{figure}

In this paper\revfivechanges{,} we propose a novel learning-based approach for estimating the optimal grasp pose to unfold cloth using the regrasping-in-the-air technique. Instead of relying on hand-crafted heuristics, we propose a deep neural network that directly predicts the grasp point from a single RGB image of the cloth suspended in the air. The network is trained in a fully supervised manner to predict the grasp pose that maximizes cloth unfolding when stretched, as defined by the annotator. \revthreechanges{In this way, deformability is handled through learning-based point supervision and a high-capacity visual backbone that can adapt grasp prediction to the currently observed cloth configuration.} Our method not only identifies the grasp location but also fully regresses a 6-DoF gripper pose, enabling precise execution of the unfolding sequence. Furthermore, by leveraging a dense prediction framework, multiple grasp candidate \revthreechanges{locations} are generated, allowing for flexible selection based on kinematic feasibility. Our approach, CeDiRNet-6DoF, builds upon the point detection capabilities of CeDiRNet~\cite{Tabernik2024PR} and extends it to predict full 6-DoF grasp poses in Euler angles as depicted in Fig.~\ref{fig:CeDiRNet-6DoF-arch}. \revfivechanges{Unlike cloth methods that typically predict only grasp points or regions and rigid-object methods that often assume stable object geometry, CeDiRNet-6DoF jointly regresses dense grasp locations and full orientation from one RGB image; depth completes the position, and kinematic-feasibility-aware selection integrates the candidates into a real-robot unfolding pipeline.}

%In this paper, we propose a novel learning-based approach for estimating the best grasping pose that unfolds the cloth with the regrasping-in-the-Air~\cite{} technique. Instead of relaying on heuristics to find the optimal point on the cloth and its grasp pose, we propose to learn a deep neural network that will estimate a point on the cloth depending on the observed configuration of the cloth being held by the first robot. The network estimates not only a point on the cloth, but also fully regresses a 6-DoF pose of the gripper performing the grasping sequence. Moreover, multiple grasp poses can be estimated using a dense prediction, thus enabling a flexible planning and execution by selecting grasp candidates with optimal joint configurations. Our proposed method, CeDiRNet-6DoF, leverages the capabilities of point detection approach CeDiRNet~\cite{Tabernik2024PR}, while extending it to regress a full 6-DoF grasp poses as Euler angles. By incorporating a regression of grasp positions alongside segmentation and Euler angles, parameterized using sine and cosine functions, our method achieves a more comprehensive and accurate representation of the grasp configuration necessary for effective cloth manipulation fully depended on the observed cloth configuration. 

Our contributions are threefold: i) we introduce CeDiRNet-6DoF, a novel deep-learning-based approach for grasp pose estimation for cloth unfolding using \revfivechanges{a} regrasping-in-the-air strategy; ii) we rigorously evaluate our approach following the ICRA 2024 competition rules, evaluating on the same competition items plus additional items, demonstrating its superior performance compared to competition methods; and iii) we provide a detailed ablation study that highlights the impact of our design choices, such as background randomization, image cropping, joint segmentation\revfivechanges{,} and the integration of RGB-D data, on the overall performance. These contributions lay the groundwork for more versatile and reliable robotic systems capable of handling the complex task of cloth manipulation. This has also been demonstrated in the ICRA 2024 cloth competition, where CeDiRNet-6DoF achieved second place and \revfivechanges{outperformed nine} of \revfivechanges{the ten competing} methods.

\section{Related Work}
\revthreechanges{

The most closely related literature concerns cloth grasp selection~\revfiveremovedcite{Caporali2020,Zhu2023}\cite{Chen2023b,Barbany2025} and in-air grasp prediction~\cite{rome_learning_2026,sunil_reactive_2025}, while rigid-object 6-DoF pose~\revfiveremovedcite{Sundermeyer2023}\cite{Liu2025a} and grasp~\cite{wu2025economic,graspcot2025,zerograsp2025} estimation provides related methodological approaches that are not always applicable to cloth manipulation due to rigid-object assumptions.

\subsection{Cloth Grasp Selection and Manipulation}

Early work on deformable-object grasping relied mainly on hand-crafted visual cues such as corners and edges for grasp-point selection~\revfiveremovedcite{Caporali2020}\cite{Maitin-Shepard2010,Doumanoglou2014}. These approaches can work well when data are scarce, but are less robust to severe self-occlusions, folds, and appearance variability common in real-world cloth manipulation.

Recent learning-based methods instead infer grasp-relevant structure directly from visual observations\revfivechanges{, including target grasp-point detection} and region-aware garment segmentation \revfivechanges{for structural-region grasp selection}~\cite{Corona2018,Chen2023b}\hspace{0.3em}\revfiveremovedcite{Qian2020,Gabas2021,Zhu2023}\unskip. \revfivechanges{Such approaches} show that cloth grasp selection can be learned effectively from image data, but typically focus on grasp points or graspable regions rather than direct 6-DoF gripper-pose prediction.

Related recent garment-manipulation systems also address broader bimanual folding or manipulation tasks~\revfiveremovedcite{Xue2023,Avigal2022}\cite{Barbany2025,Deng2024,Huang2024a}. However, such systems often predict coordinated grasp pairs, use richer task-level supervision, or target broader manipulation policies rather than direct single-grasp 6-DoF prediction for unfolding.

\subsection{In-air Cloth Grasping and Direct Grasp Prediction}

Among cloth-specific methods, an important distinction is between tabletop manipulation and in-air manipulation. Prior work on deformable-object grasp-point selection spans both tabletop grasping~\cite{huang_task_2023,chen_deformpam_2025,ha_flingbot_2021} and in-air manipulation~\cite{sunil_visuotactile_2022,sunil_reactive_2025,rome_learning_2026}. Tabletop methods often select grasp points for cloth lying on a surface, using task-oriented metrics~\cite{huang_task_2023}, preference- and reward-guided action selection~\cite{chen_deformpam_2025}, or end-to-end policies for dynamic bimanual primitives~\cite{ha_flingbot_2021}. While highly relevant to deformable manipulation, these methods do not address direct 6-DoF in-air grasp prediction and often rely on tabletop-specific heuristics or predefined primitives for executing grasps.

On the other hand, the more closely related in-air approaches typically rely on stronger sensing or observation assumptions than our setting. Examples include visuotactile edge tracing~\cite{sunil_visuotactile_2022}, confidence-aware visuotactile affordance prediction~\cite{sunil_reactive_2025}, and dual-view policies trained extensively in simulation~\cite{rome_learning_2026}. These methods therefore require tactile sensing, multiple viewpoints, or broader policy-level assumptions that differ materially from the single-view benchmark considered here. This motivates our positioning of CeDiRNet-6DoF as a single-view method for direct 6-DoF grasp prediction in a full real-robot unfolding pipeline.

\subsection{6-DoF Pose Estimation and Grasping for Rigid Objects}

Deep-learning-based 6-DoF pose estimation from a single RGB or RGB-D image is well established for rigid objects, including keypoint-correspondence and voting methods~\revfiveremovedcite{Yu2020}\cite{Peng2019}, dense correspondence or coordinate-map regression~\cite{Li2019CDPN:Estimation}\revfiveremovedcite{Hu2019b,Park2019a,Zakharov2019}\unskip, direct rotation prediction \revfivechanges{and continuous angular regression}~\revfiveremovedcite{Bukschat2020}\cite{Xiang,Beyer2015}, and \revfivechanges{geometry-guided pose estimation}~\cite{Liu2025a}\revfiveremovedcite{Wang2021d,Di2021,Sundermeyer2023}\unskip. These methods estimate object pose relative to a reference shape or canonical object frame, which is natural for rigid objects but much less applicable to deformable cloth, where configuration changes continuously and cannot be mapped reliably by a single rigid transformation. This motivates moving from 6-DoF pose estimation to methods that predict executable grasps directly.

Direct 6-DoF grasp estimation for rigid objects has likewise progressed from pose-plus-grasp pipelines~\revfiveremovedcite{Wang2023e}\cite{Cong2023} to end-to-end grasp prediction from RGB-D or point-cloud observations~\cite{Shang2022,Cong2023}. Recent examples include Generalizing-Grasp~\cite{generalizinggrasp2024} for domain-prior-guided generalization, Language-Driven 6-DoF Grasp Detection~\cite{negativepromptguidance2024} for language-conditioned diffusion-based grasp generation, Region-aware Normalized Grasp Space~\cite{regionnormalizedgrasp2024} for efficient single-view detection, ZeroGrasp~\cite{zerograsp2025} for reconstruction-enabled zero-shot grasping, and GraspCoT~\cite{graspcot2025} for language- and reasoning-augmented prediction. Among them, EconomicGrasp~\cite{wu2025economic} is especially relevant in our setting because it remains comparatively practical while still performing direct single-view 6-DoF grasp prediction, learning from point clouds through dense objectness and graspness prediction with sparse supervision.

These rigid-object methods are therefore useful comparison points, but transferring them to hanging cloth remains nontrivial because they are often designed around rigid-object assumptions that do not transfer naturally to deformable garments, including grasp definitions tied to stable object geometry or to grasping the object as a whole rather than selecting a small locally suitable grasp region. In addition, several recent methods do not yet provide straightforward public reproducibility for train-and-evaluate use in our benchmark.
}
\subsection{Benchmarks and Datasets}

Progress in learning-based cloth grasping has also been driven by larger datasets and more structured evaluation protocols. Lips et al. \revfivechanges{demonstrated learning garment keypoints from} synthetic \revfivechanges{data} with semantic keypoint annotations~\revfiveremovedcite{Lips2022}\cite{Lips2024LearningData}. Our prior work introduced the ViCoS Towel Dataset together with CeDiRNet-3DoF~\cite{Tabernik2024RAL}, enabling dense learning-based grasp-point detection under cloth-specific supervision, but still focusing on 3-DoF prediction and image-level evaluation. A benchmark for full 6-DoF evaluation with the regrasping-in-the-air strategy was then introduced through the ICRA 2024 Cloth Competition~\cite{DeGusseme2024BenchmarkingGrasp}, providing a training dataset, an execution setup, and an evaluation protocol for end-to-end robotic cloth unfolding.

\subsection{Positioning of CeDiRNet-6DoF}

Against this backdrop, we position CeDiRNet-6DoF in the regrasping-in-the-air setting of the ICRA 2024 benchmark. \revthreechanges{In contrast to tabletop cloth-grasping methods, we target direct in-air 6-DoF grasp prediction. In contrast to tactile or multi-view in-air approaches, our method uses a single visual observation. In contrast to broader bimanual garment-manipulation systems~\revfiveremovedcite{Xue2023,Avigal2022}\cite{Barbany2025,Deng2024,Huang2024a}, we focus on predicting a single executable grasp pose aligned with the benchmark setup rather than coordinated grasp pairs or broader task policies.

Compared to generic rigid-object 6-DoF grasp estimators such as the point-cloud-based EconomicGrasp~\cite{wu2025economic}, our formulation eliminates rigid-object assumptions and learns from single grasp-point annotations, without requiring explicit objectness, segmentation, or grasp definitions tied to the object as a whole. It also estimates the grasp pose from RGB images rather than point clouds, with depth used only for positioning in 3D. Rather than relying on a stable canonical object pose~\cite{Liu2025a}\revfiveremovedcite{Wang2021d}\unskip{} or an explicit model of cloth deformation~\cite{rome_learning_2026}, the method learns grasp selection from supervision tied to the current cloth state. This is particularly important for cloth, where the geometry changes continuously and grasp success depends strongly on local visual cues such as corners, folds, edges, and the hanging geometry. Point-based supervision therefore provides a natural way to adapt direct grasp prediction to deformable-object manipulation.

Building on our prior CeDiRNet-3DoF framework~\cite{Tabernik2024PR}, CeDiRNet-6DoF retains the dense point-localization formulation while extending it to full 6-DoF grasp-pose prediction and integrating it into a complete real-robot unfolding pipeline. While the backbone architecture remains the same as in CeDiRNet-3DoF, using ConvNeXt with an FPN, CeDiRNet-6DoF introduces 6-DoF-specific adaptations, including orientation regression, segmentation supervision, background randomization, and robot-aware cropping.}

\section{Method}

In this section, we describe our proposed approach for full 6-DoF grasp pose estimation using CeDiRNet-6DoF and its integration into a robotic cloth manipulation pipeline. The method builds upon our previous work on cloth key-point detection in 3-DoF~\cite{Tabernik2024RAL}, extending it to predict full 6-DoF grasp poses. {We build on top of definitions and notations introduced in CeDiRNet~\cite{Tabernik2024PR} and its 3DoF variant~\cite{Tabernik2024RAL}, but provide basic definitions here as well.} We also describe our planning and grasp execution strategies that perform the final stretching of the grasped cloth.

\subsection{CeDiRNet}
Our proposed network builds on \revfivechanges{the} Center Direction Regression Network (CeDiRNet) for point detection~\cite{Tabernik2024PR}. Here, we provide a quick formulation, while the reader is referred to~\cite{Tabernik2024PR,Tabernik2024RAL} for more details. CeDiRNet performs point detection using a two-stage approach: i) a dense regression of center vectors in the first stage, and ii) a localization using a lightweight, domain-agnostic network in the second stage. In the first stage\revfivechanges{,} given \revfivechanges{an} input image $\mathcal{I}$, a dense regression network outputs two dense fields, $C_{\sin}, C_{\cos} =  \mathrm{DenseRegNet}(\mathcal{I})$, which encode the direction $\phi$ toward the target location in the image plane for every image pixel. The direction to target $\phi$ is parameterized with a trigonometric function. $\mathrm{DenseRegNet}$ can also output a third dense output field $R$ that is regressed to \revfivechanges{the logarithm} of \revfivechanges{the} distance $r$ to the closest target for each pixel. In the second stage, a $\mathrm{LocNet}$ is used to extract the exact point locations from the regressed center-directions fields, ${\mathcal{O}_{\mathrm{cent}} = \mathrm{LocNet}(C_{\sin}, C_{\cos})}$, where $\mathcal{O}_{\mathrm{cent}}$ is a probability map for target location with local maxima representing final predicted locations.

\subsection{Grasp Pose Estimation in 6-DoF}

We extend CeDiRNet for 6-DoF pose \revfivechanges{regression, following the} 3-DoF \revfivechanges{extension in}~\cite{Tabernik2024RAL}, but now regress a three-dimensional orientation instead of a single-dimensional one{, while \revfivechanges{also} regressing cloth segmentation}. \revthreechanges{Concretely, the CeDiRNet-3DoF variant regressed a single sin/cos orientation pair, whereas CeDiRNet-6DoF uses three parallel orientation heads, one for each Euler-angle component, with each head regressing its own sin/cos pair. This results in a CeDiRNet-6DoF model.}

\revfivechanges{We} define a 3-dimensional location $X \in \mathbb{R}^3$ with Euler angles $\theta \in \mathbb{R}^3$ as a grasp pose relative to the camera frame. The dense regression network in CeDiRNet-6DoF now predicts the following fields:
\begin{equation}
C_{\sin/\cos}, (D^x,D^y,D^z)_{\sin/\cos}, S, R = \mathrm{DenseRegNet}(\mathcal{I}),
\end{equation}
where $C_{\sin}$ and $C_{\cos}$ encode the center-directions $\phi$ toward the closest grasp points projected from the 3D position $X \in \mathbb{R}^3$ into the image plane $x \in \mathbb{R}^2$, while $D_{\sin}$ and $D_{\cos}$ encode corresponding \revfivechanges{three-dimensional} Euler angles $\theta$ \revfivechanges{about the} x\revfivechanges{-}, y\revfivechanges{-,} and \revfivechanges{z-axes} (roll, pitch, and yaw components) relative to the camera frame. Additionally, $S$ is the \revfivechanges{cloth-segmentation mask} and $R$ is \revfivechanges{the} log-distance $r$ to the closest grasp point. The dense regression network predicts one set of Euler angles for every pixel location, allowing for the estimation of multiple grasp points in the same image. \revthreechanges{Because the dense prediction and localization stages can return multiple local maxima, the system produces a set of candidate grasp locations from a single image, and each candidate location is paired with one regressed 6-DoF pose.} Following~\cite{Tabernik2024PR} and~\cite{Tabernik2024RAL}, the Euler angles are reparametrized with the trigonometric function to convert values from $[-\infty,\infty]$ into $[-1,1]$, while also removing angle discontinuities at $2\pi$. The original Euler angles for roll (x-axis), pitch (y-axis)\revfivechanges{,} and yaw (z-axis) are recovered element-wise as:

\begin{equation}
\boldsymbol{\theta}
= \operatorname{atan2}(\mathbf{D}_{\sin},\mathbf{D}_{\cos}),
\end{equation}

where $\mathbf{D}_{\sin}=(D^x_{\sin},D^y_{\sin},D^z_{\sin})$
and $\mathbf{D}_{\cos}=(D^x_{\cos},D^y_{\cos},D^z_{\cos})$.
%\begin{equation}
%\begin{aligned}
%\theta &= \left[\tan^{-1}\left(\frac{D^x_{\cos}}{D^x_{\sin}}\right), \tan^{-1}\left(\frac{D^y_{\cos}}{D^y_{\sin}}\right),  \tan^{-1}\left(\frac{D^z_{\cos}}{D^z_{\sin}}\right)\right].
%\end{aligned}
%\end{equation}

The final 6-DoF grasp pose prediction is then constructed by combining the predicted 2D \revfivechanges{grasp-pose} location $\hat{x}$ retrieved by the $\mathrm{LocNet}$ stage and the predicted Euler angles $\hat{\theta}(\hat{x})$ read at location $\hat{x}$. The 3D position of the grasp point is recovered by combining the pixel location with the depth measurement provided by an RGB-D sensor.

\paragraph*{Euler-angle representation}
\revthreechanges{We use Euler angles mainly as a practical representation choice for this benchmark and architecture. The training annotations in the ICRA 2024 cloth competition dataset are already provided in Euler-angle form, so keeping the same target representation preserves direct alignment between labels and regression targets. The sine/cosine parameterization mitigates angular wrap-around discontinuities, while remaining compact for dense per-pixel prediction. We note, however, that Euler angles still have known limitations, such as coupling between axes and sensitivity near singular configurations. Alternative representations, such as quaternions and rotation matrices, may mitigate some of those limitations, but come with other trade-offs. For instance, quaternions have sign-equivalence issues while rotation matrices would require higher-dimensional dense outputs together with additional structural consistency constraints. We therefore leave exploration of alternative representations for future work.}

\paragraph*{Backbone and dense regression network}
CeDiRNet is agnostic to the specific backbone or dense regression network. The backbone architecture for our 6-DoF version remained the same as in our prior 3-DoF variant~\cite{Tabernik2024RAL}, employing a ConvNeXt backbone with a Feature Pyramid Network (FPN) for multi-scale feature aggregation. This backbone replaces the ResNet backbone originally used in CeDiRNet~\cite{Tabernik2024PR}, while the localization network remained identical to the originally proposed one. \revthreechanges{In the context of cloth manipulation, retaining a generic high-capacity visual backbone together with dense prediction enables the model to learn grasp-relevant regularities directly from supervision. This design is well matched to severe self-occlusion, configuration-dependent geometry, and substantial appearance variability, where rigid hand-crafted assumptions are difficult to maintain.}

\subsection{\revfourchanges{Data Input}}
\revfourchanges{We crop \revfivechanges{the} input image based on \revfivechanges{the robots'} workspace positions to suppress the background and focus on cloth details. Since our setup keeps the cloth between the two robots, we use the robot-base positions to define the left, right, and lower crop boundaries. The upper boundary is placed 5~cm above the gripper of the left robot, which holds the cloth (see Fig.~\ref{fig:setup-and-detection}). We apply the same crop during training and inference, and the resulting cropped image is used as input to the final model. By default we use RGB input, but for the RGB-D variants we modify only the input layer by adding depth: distance to the world XZ-plane through the holding gripper and perpendicular to the ground, normalized from $[-50,50]$~cm to $[-1,1]$. This object-centric encoding avoids absolute camera depth and reduces camera-position/viewpoint dependence.}

\subsection{Learning}
The model is trained in a fully supervised manner using an annotated 6-DoF grasp pose for each training sample. The dataset and annotations we used are described in Section~\ref{sec:dataset}. Training of $\mathrm{DenseRegNet}$ \revfivechanges{follows}~\cite{Tabernik2024RAL}, \revfivechanges{with} all dense output fields \revfivechanges{trained jointly}. \revthreechanges{We use a shared dense output head for $C+R+S$, together with three parallel orientation heads for $D^x$, $D^y$, and $D^z$, resulting in a total of four individual heads. Each orientation head predicts the sin/cos parameterization for one Euler axis. This enables specialization for center directions with distance and segmentation within the shared dense prediction head, as well as specialization for individual Euler axes.} With the exception of weights for the final dense output heads, all other encoder and decoder weights are shared. Losses for each individual output head are balanced using uncertainty weighting~\cite{Kendall2018} { as already proposed in \revfivechanges{CeDiRNet-3DoF}~\cite{Tabernik2024RAL}}:
\begin{align}
    \mathcal{U}_w(\hat{\mathcal{L}},\sigma) =& \frac{1}{2\sigma^2} \hat{\mathcal{L}} +  \log\sigma,
\end{align}
%\begin{align}
%    \hat{\mathcal{L}}_{\phi} =& \frac{1}{2\sigma_{\phi}^2} \mathcal{L}_\phi +  \log\sigma_{\phi},\\
%    \hat{\mathcal{L}}_{\theta_x} =&\frac{1}{2\sigma_{\theta_x}^2} \mathcal{L}_{\theta_x} + \log\sigma_{\theta_x}, \\ 
%    \hat{\mathcal{L}}_{\theta_y} =&\frac{1}{2\sigma_{\theta_y}^2} \mathcal{L}_{\theta_y} + \log\sigma_{\theta_y}, \\
%    \hat{\mathcal{L}}_{\theta_z} =&\frac{1}{2\sigma_{\theta_z}^2} \mathcal{L}_{\theta_z} + \log\sigma_{\theta_z}, \\
%    \hat{\mathcal{L}}_{seg} =&\frac{1}{2\sigma_{seg}^2} \mathcal{L}_{seg} + \log\sigma_{seg},
%\end{align}
\revfivechanges{The} final loss \revfivechanges{is} composed of the individual losses:

\begin{align}
\mathcal{L} =~& \mathcal{U}_w(\hat{\mathcal{L}}_{\phi} + \hat{\mathcal{L}}_{r},\sigma_{\phi}) + 
                \mathcal{U}_w(\hat{\mathcal{L}}_{seg}, \sigma_{seg})~+ \nonumber \\
            &  \mathcal{U}_w(\hat{\mathcal{L}}_{\theta_{x}}, \sigma_{\theta_{x}}) + 
                \mathcal{U}_w(\hat{\mathcal{L}}_{\theta_{y}}, \sigma_{\theta_{y}}) + 
                \mathcal{U}_w(\hat{\mathcal{L}}_{\theta_{z}}, \sigma_{\theta_{z}}) 
\end{align}
with the weighting values $\sigma_{\phi}$, $\sigma_{\theta_{x}}$, $\sigma_{\theta_{y}}$, $\sigma_{\theta_{z}}$ and $\sigma_{seg}$ acting as additional learnable parameters defining the uncertainty weight for each corresponding loss. This eliminates the need for manually setting additional hyperparameter weights and prevents gradients f\revfivechanges{r}o\revfivechanges{m} one task from \revfivechanges{dominating and} negatively influencing the other tasks {as demonstrated in \revfivechanges{our} prior 3-DoF version~\cite{Tabernik2024RAL}}. Segmentation $S$ and log-distance $R$ are learned during training through $\hat{\mathcal{L}}_{seg}$ and $\hat{\mathcal{L}}_{r}$ losses, but are not needed for inference; \revtwochanges{instead, they provide auxiliary-task supervision during training and can have regularization-like effects on the shared representation.} Note that log-distance $R$ was already used in CeDiRNet-3DoF~\cite{Tabernik2024RAL}, although not officially reported in the paper.

We use binary cross-entropy for segmentation loss { $\hat{\mathcal{L}}_{seg}$ } and, similar to~\cite{Tabernik2024RAL}, L1 for all other regression losses, { where $\hat{\mathcal{L}}_\theta$ is applied to $(D_{\sin}, D_{\cos})$ fields for roll, pitch and yaw; $\hat{\mathcal{L}}_{\phi}$ is applied to $(C_{\sin}, C_{\cos})$ for center-directions and $\hat{\mathcal{L}}_{r}$ is applied to log-distance $R$.} All regression losses are computed only for foreground regions ($30\times30$ pixels around each grasp point), as in~\cite{Tabernik2024PR,Tabernik2024RAL}. Note that we ignore the background only during training; during inference, we regress for all pixels but read the final pose orientation only at the detected grasp point location.

\subsection{\revfourchanges{Inference}}

\revfourchanges{Starting from the cropped robot-workspace image in RGB or RGB-D, DenseRegNet jointly predicts center directions and three orientation fields, as well as auxiliary training outputs $S$ and $R$, which are discarded during inference. LocNet extracts 2D candidates whose corresponding orientation-field values provide roll, pitch, and yaw, while camera depth then provides the 3D position for both RGB and RGB-D variants. Before planning, 2.5~cm is added along the grasp-pose z-axis to slightly increase the depth at which the gripper executes the grasp.}

\subsection{Planning and Grasping}\label{sec:planning_and_grasping}
Once the 6-DoF grasp pose is estimated by CeDiRNet-6DoF, this pose must be transformed into executable robot motions. First, the grasp pose defined in the camera frame is converted into the robot's world frame using the known extrinsic calibration between the RGB-D sensor and the robot base. Given that CeDiRNet-6DoF detects multiple potential grasp poses, we sort them by \revfivechanges{detection} confidence based on \revfivechanges{the} $\mathcal{O}_{\mathrm{cent}}$ probability value from $\mathrm{LocNet}$. We also filter out detections that are further away than the maximal reach of the robot arm. \revthreechanges{At inference time, the candidate poses are therefore ranked by detection confidence and filtered by reachability, inverse-kinematics validity, and trajectory feasibility before one executable grasp is selected.} Then, we select a grasp pose with the highest confidence score and a valid \revfivechanges{inverse-kinematics} solution for the final execution.

An analytical inverse kinematics solver~\cite{anaIK} is applied to map the target 6-DoF pose into the corresponding joint space configuration. Given that our robotic manipulators have more than six degrees of freedom, multiple inverse kinematics solutions can exist for a single target pose. To resolve this redundancy, our system iterates over potential last joint configurations and selects the one closest—by Euclidean distance—to empirically defined desirable configurations. This ensures that the computed joint configuration is both feasible and efficient for execution.

\begin{algorithm}
\caption{Robust 6-DoF Grasp Pose Execution}
\label{alg:grasping}
\begin{algorithmic}[1]
\Require current joint position $\mathbf{q}_{\text{right}}$ for the right arm performing the grasp; a predefined set of joint configurations $Q_{\text{left}} = \{\mathbf{q}_{\text{default}},\, \mathbf{q}_{\text{near}},\, \mathbf{q}_{\text{far}}\}$ for the left arm, which holds the cloth and defines its position; maximum reach $d_{\text{max}}$; and extrinsic calibration $R$.

\Ensure Successful grasp execution or report failure

\For{each $\mathbf{q}_i \in Q_{\mathrm{left}}$}
    \State \textbf{Reposition} cloth by setting left arm joints to $\mathbf{q}_i$
    \State \Comment{\textit{Detect grasp pose candidates in camera frame}}
    \State $P \gets \mathrm{CeDiRNet6DoF}(\mathcal{I} | \mathbf{q}_i)$ sorted by $\mathcal{O}_{\mathrm{cent}}$
    \State \Comment{\textit{Project $p$ to robot frame and filter out-of-reach}}
    \State $P_w \gets \{\, R \cdot p | p \in P \text{ and } \|R \cdot p\| \le d_{\mathrm{max}} \,\}$ 
    \If{$P_w$ from $\mathbf{q}_{\mathrm{default}}$ was empty}
        \State \Return \textbf{failure} - no grasp poses
    \EndIf
    \State \Comment{\textit{Compute all inverse kinematics for all candidates}}
    \State $S_w \gets [\mathrm{ComputeIK}(p_w) \mid p_w \in P_w]$
    \State $s \gets $ select $s \in S_w $ within joint limits, no collision, and highest $\mathcal{O}_{\mathrm{cent}}$
    \If{$s$ exists}
        \State $T_{\mathrm{traj}} \gets \mathrm{PlanTrajectory}(\mathbf{q}_{\mathrm{right}}, s)$
        \State \textbf{Execute} $T_{\mathrm{traj}}$ with right arm
        \State \Return \textbf{success}
    \EndIf
\EndFor
\State \Return \textbf{failure}
\end{algorithmic}
\end{algorithm}

For the grasp execution, a motion planning module~\cite{sucan2012the-open-motion-planning-library} computes a collision-free trajectory from the robot's current configuration to the target configuration, calculated from the target grasp pose using inverse kinematics. We employ an efficient planner, specifically the RRT-Connect~\cite{kuffner2000}, to explore the high-dimensional configuration space and generate a viable path, while real-time collision checking is performed against a model of the environment—including both robot arms, the table, the virtual boundary walls and the detected cloth.

\begin{figure}
    \centering
    \includegraphics[width=\linewidth]{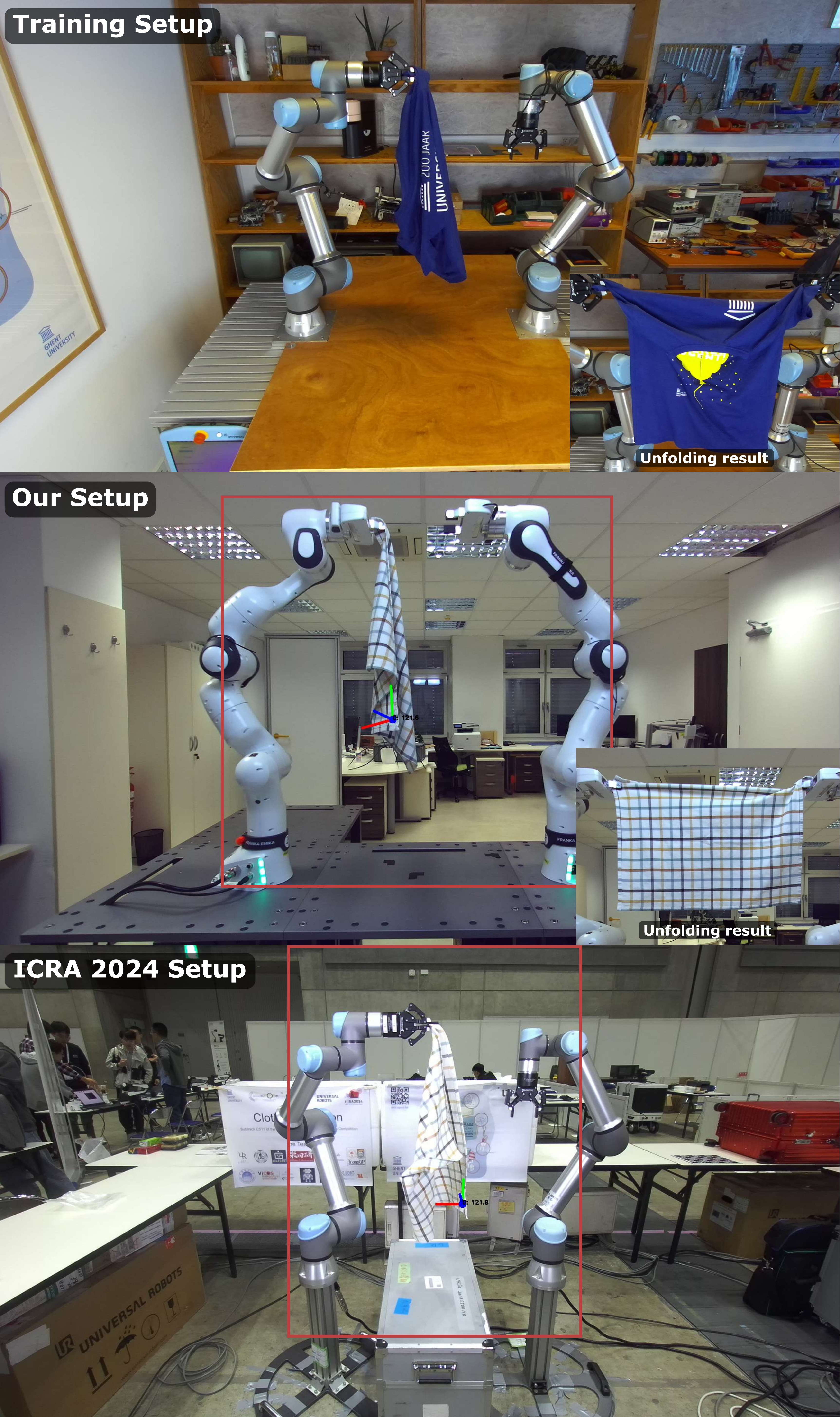}
    \caption{Captured images from three different setups: (i) the competition training dataset provided by the organizers (top row), showing the ground truth data; (ii) our setup (middle row), showing the inference results; and (iii) the competition setup at ICRA 2024 (bottom row), showing the inference results from the competition. The red bounding boxes indicate the cropping boundaries based on the robot's position, as used in our approach. }
    \label{fig:setup-and-detection}
\end{figure}

In scenarios where all grasp poses within \revfivechanges{the} potentially reachable space \revfivechanges{lack a} valid arm joint solution or pose a risk of collision, the system adopts a heuristic re-planning strategy as detailed in Algorithm~\ref{alg:grasping}. This involves a controlled repositioning of the robot arm holding the cloth followed by a re-evaluation of the grasp pose, thereby ensuring robust execution even under challenging conditions. We define an initial joint configuration $\mathbf{q}_{\text{default}}$ and two secondary joint configurations, $\mathbf{q}_{\text{far}}$ and $\mathbf{q}_{\text{near}}$, each resulting in \revfivechanges{a} different gripper position \revfivechanges{for holding} the cloth. If no grasp poses are found within \revfivechanges{the} potentially reachable space for the initial cloth position, no re-planning is performed, \revfivechanges{and we consider} this case a failed grasp. By executing a slow move during re-positioning, the textile hardly deform\revfivechanges{s}. The initial joints $\mathbf{q}_{\text{default}}$ position the gripper 10 cm from the center-point between the robots towards the left robot. In the second attempt, the gripper was positioned 20 cm toward the right robot for $\mathbf{q}_{\text{near}}$ to bring it closer to the grasp arm, while in the last attempt, we positioned the cloth 10 cm from the initial hanging position in the other direction for $\mathbf{q}_{\text{far}}$, to move it further away from the grasp arm.

Overall, this integrated planning and grasping framework enables robust and efficient robotic manipulation of deformable cloth objects by ensuring that the selected 6-DoF grasp poses are reliably translated into safe, executable motions.

\section{Experiments}
In this section\revfivechanges{,} we extensively evaluate the whole pipeline for the robotic dual-arm cloth unfolding task using the regrasping-in-the-air strategy~\cite{Maitin-Shepard2010,Doumanoglou2014,DeGusseme2024BenchmarkingGrasp}.

\subsection{Task}\label{sec:task}

We follow the task definition from the ICRA 2024 cloth competition~\cite{DeGusseme2024BenchmarkingGrasp} that involves unfolding the cloth through the following steps: 
\begin{enumerate}[label=Step \roman*), leftmargin=*]
    \item \revfivechanges{A} cloth randomly positioned on the table is picked up with the right robot arm at the highest point and lifted in the air.
    \item The lowest point on the cloth hanging in the air is then \revfivechanges{used} as the first grasp point. The cloth is grasped at this point and lifted into the initial configuration with the left arm for the next grasp.
    \item The optimal second grasp pose for fully unfolding the cloth must be estimated and executed with the right robot arm.
    \item Finally, the cloth is completely unfolded by stretching it in the air in front of the camera with both robot arms maintaining a grip at the grasp points. 
\end{enumerate}
%Note that the process can start with either right or left hand, but we used right hand to follow the ICRA competition process. Steps (i),(ii), and (iv) are predefined, while step (iii) is the most crucial and must be performed autonomously in the optimal way.  The success is measured by the surface area of the unfolded cloth after the stretching in the air relative to the reference surface area at the manually defined optimal stretch position with the largest surface. Success is measured by comparing the surface area of the unfolded cloth after it has been stretched in the air to the reference surface area at the manually defined optimal stretch position, where the surface is maximized. More details on surface area measurement are provided in Section~\ref{sec:evaluation-protocol} on evaluation metrics.

Note that the process can be initiated with either the right or left hand; however, we use the right hand to align with the procedure followed in the ICRA 2024 competition, \revfivechanges{which results} in the left arm \revfivechanges{holding} the cloth in the air and the right arm \revfivechanges{performing} the grasp. Steps (i), (ii), and (iv) are predefined, while step (iii)—grasp selection—is the most critical and must be executed autonomously to ensure optimal unfolding. Success is measured by comparing the surface area of the cloth after in-air stretching to a reference surface area obtained at a manually defined optimal stretch configuration, where the surface coverage of the cloth is maximized. Additional details on surface area computation are provided in Section~\ref{sec:evaluation-protocol}.

\subsection{Robot Setup} 
Our setup closely replicates the ICRA 2024 cloth competition setup, utilizing a dual-arm configuration with the robots positioned 90 cm apart. The same RGB stereo-depth camera, the Zed2i RGB-D camera with passive stereo, was mounted behind the robots at a slight downward angle. While the competition employed two UR5e robots equipped with wrist-mounted force-torque sensors and Robotiq 2F-85 grippers, we used two 7-DoF Franka Robotics FR3 robots with basic Franka Hand grippers featuring plastic fingertips. Since high force feedback accuracy was not required, we extrapolated forces from the motors. This method was sufficient for reliable use during the final stretching phase. All three setups (ours, the \revfivechanges{training-data setup,} and the ICRA 2024 competition \revfivechanges{setup}) are depicted in Fig.~\ref{fig:setup-and-detection}.

While both robots have a similar reach, 855 mm for \revfivechanges{the} FR3 and 850 mm \revfivechanges{for the} UR5e, they differ in the \revfivechanges{number of degrees of freedom}, \revfivechanges{seven} for \revfivechanges{the} FR3 and \revfivechanges{six} for \revfivechanges{the} UR5e, and the range of joint motion, with \revfivechanges{a larger} range for \revfivechanges{the} UR5e \revfivechanges{than} t\revfivechanges{he} FR3, resulting in \revfivechanges{a} differently shaped workspace \revfivechanges{f}o\revfivechanges{r} the UR5e. We thus included a \revfivechanges{simple} error-handling procedure to reposition the cloth closer \revfivechanges{to} or further away from the left robot arm as described in Section~\ref{sec:planning_and_grasping}.

\subsection{Training Datasets} \label{sec:dataset}

We used two datasets for training: i) \revfivechanges{the} ViCoS Towel Dataset~\cite{Tabernik2024RAL} extended with additional samples for pre-training, and ii) a dataset provided by the organizers of the ICRA 2024 competition~\cite{de_gusseme_2025_14621179} for fine-tuning the model for the unfolding task. We refer to the latter as the competition dataset.

\paragraph{The ViCoS Towel Dataset} The ViCoS Towel Dataset~\cite{Tabernik2024RAL} consists of 8,000 real and 12,000 synthetic RGB-D images of towels lying flat on the table, captured from a top-down view and annotated with segmentation, corner keypoints, and 1D orientation for a 3-DoF grasp pose. We extended this dataset with an additional 559 samples of the same towels under more difficult conditions. { The same setup used to capture the original 8,000 real images was employed to collect 459 additional samples, while the remaining 100 samples were acquired using an RGB‑only camera mounted in a top‑down view at a resolution of $1341 \times 1012$. To increase the difficulty of the dataset, we focused primarily on additional folded and crumpled configurations, which have proven challenging in the CeDiRNet‑3DoF evaluation. These extra images boosted the F1 score on the ViCoS Towel 3‑DoF benchmark from 78.0~\% to 79.8~\%. We have also made the expanded image set publicly available\footnote{\url{https://go.vicos.si/toweldatasetextra}}.} We used real and synthetic training data from the ViCoS Towel Dataset and an additional 559 samples to pre-train our model.

\paragraph{The ICRA 2024 cloth competition dataset} To fine-tune our model for the unfolding task, we used the ICRA 2024 cloth competition dataset~\cite{de_gusseme_2025_14621179} {as described in~\cite{DeGusseme2024BenchmarkingGrasp}. The dataset consists} of 503 samples captured from 28 items (13 towels and 15 T-shirts) \revfivechanges{using} the competition setup. An example image is depicted in the top row in Fig.~\ref{fig:setup-and-detection}. Each sample contains a single grasp point annotated \revfivechanges{with a} 6-DoF \revfivechanges{pose} by a human operator, with the camera, robot arm base, and gripper positions provided in world coordinates. \revthreechanges{This annotation format therefore provides one target grasp pose per image, which also steered our formulation toward predicting one 6-DoF pose for each detected grasp candidate.} The dataset also includes images \revfivechanges{acquired} after \revfivechanges{execution of} the annotated \revfivechanges{grasp;} we added \revfivechanges{cloth-segmentation masks}. We split the dataset into \revfivechanges{training} and validation sets, using two T-shirts and one towel for validation \revfivechanges{and} the remaining \revfivechanges{items} for \revfivechanges{training}.

\subsection{Training and Inference Details}

We trained the grasp pose estimation model in two stages. First, we pre-trained it for detecting cloth corners, \revfivechanges{and} then we fine-tuned it for 6-DoF grasp estimation. The pre-training stage \revfivechanges{primed the model} for the visual appearance of textiles and clothes. We use\revfivechanges{d} the extended ViCoS Towel Dataset for pre-training. Since the dataset contains only 3-DoF poses, we train\revfivechanges{ed} only center-direction vectors and a single-dimensional orientation, i.e., CeDiRNet-3DoF~\cite{Tabernik2024RAL}. \revfivechanges{We} first \revfivechanges{pre-trained} with 12,000 synthetic \revfivechanges{images} and then with 5,679 real images from the ViCoS Towel Dataset training set plus our additional 559 samples. { Before starting our pre-training stage, we also initialized the ConvNeXt backbone with publicly available ImageNet pre-trained weights.} In the second stage, we fine-tune\revfivechanges{d} the full 6-DoF pose estimation model based on the pre-trained 3-DoF model weights. Since the two additional Euler-angle heads and the segmentation output \revfivechanges{we}re not part of the 3-DoF model, they \revfivechanges{we}re initialized from scratch during fine-tuning. The model \revfivechanges{wa}s fine-tuned on our training split of the ICRA 2024 competition data for 6-DoF pose estimation.

We used the same pre-training optimization process \revfivechanges{as in}~\cite{Tabernik2024RAL}, with slight modifications: increasing the epoch size to 50, resizing real images to $960\times544$ pixels, and replacing Gaussian blur augmentation with random horizontal/vertical flips and rotations. { Augmentations were performed online during training, each independently activated with a probability of $0.5$, with rotations using discrete angles at $10\degree$ increments selected from a uniform random distribution.} We used the ConvNeXt-B backbone in all our experiments. {Together with the extra 559 training images, the changes to the pre-training protocol improved the F1 score on the ViCoS Towel 3-DoF benchmark from 78.4\% to 83.3\%.} For fine-tuning on the competition dataset, we trained the model for 100 epochs using a batch size of 4, a learning rate of $10^{-4}$ (Adam optimizer), and a polynomial decay (power of 0.9). Training images were resized to $768\times768$ pixels. We performed color jittering { with a probability of $0.5$ using randomly selected values from $[-0.3,0.3]$ for brightness, contrast, saturation\revfivechanges{,} and hue; image scaling with \revfivechanges{a} scale factor \revfivechanges{uniformly sampled} from $[0.8,1.2]$;} and background randomization for data augmentation. For background randomization, we used COCO images \revfivechanges{with} two \revfivechanges{background-replacement} techniques: i) using cloth segmentation masks or ii) using depth larger than 1.2 m. We performed the former with a probability of 0.2 and the latter with 0.8.

\begin{figure*}
    \includegraphics[width=1.0\linewidth]{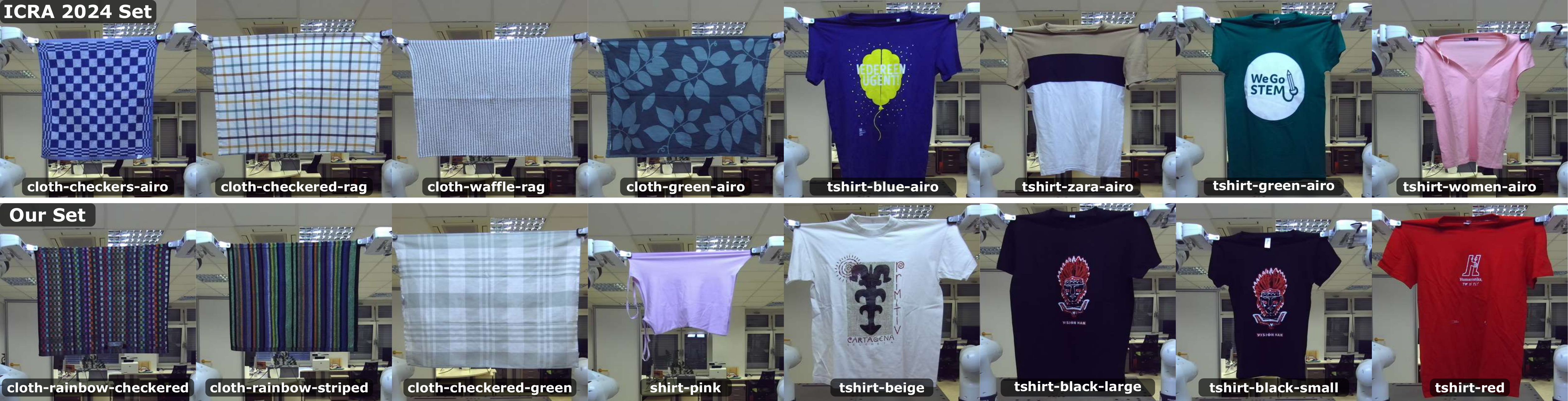}
    \caption{Cloth evaluation reference items for ICRA 2024 set (top row) and our set (bottom row).}
    \label{fig:cloth-references}
\end{figure*}

\begin{table}[h!]
\centering
\caption{Evaluation results for ICRA 2024 Set and Our Set with reported coverage, grasp success and coverage for only successful grasps. {Combined results from RUNS A and RUNS B \revfivechanges{are reported}.} Items observed during training are marked with (*).}
\label{tab:results_by_item}
\begin{tabu}{l|ccc}
\toprule
\textit{\textbf{Cloth}} & \textit{Coverage} & \textit{\makecell{Grasp\\Success}} & \textit{\makecell{Coverage for\\Successful Grasp}} \\
\toprule
\textbf{ICRA 2024 Set} & & & \\
cloth-checkered-rag* & 0.648 & 0.70 & 0.804 \\
cloth-waffle-rag* & 0.560 & 0.50 & 0.793 \\
tshirt-blue-airo* & 0.618 & 0.80 & 0.691 \\
tshirt-green-airo* & 0.667 & 0.70 & 0.808 \\
tshirt-zara-airo & 0.639 & 1.00 & 0.639 \\
tshirt-women-airo & 0.752 & 0.80 & 0.860 \\
cloth-green-airo & 0.462 & 0.60 & 0.553 \\
cloth-checkers-airo & 0.561 & 0.70 & 0.660 \\
\midrule
\rowfont{}
\textit{Average (only *)} & 0.623 & 0.675 & 0.774 \\
\rowfont{}
\textit{Average (only without *)} & 0.604 & 0.775 & 0.678 \\
\textit{Average} & 0.613 & 0.725 & 0.726 \\

\midrule
\textbf{Our Set} & & & \\
cloth-checkered-green & 0.359 & 0.30 & 0.607 \\
cloth-rainbow-checkered & 0.503 & 0.70 & 0.564 \\
cloth-rainbow-striped & 0.455 & 0.30 & 0.636 \\
shirt-pink & 0.704 & 1.00 & 0.704 \\
tshirt-beige & 0.738 & 0.90 & 0.769 \\
tshirt-black-large & 0.525 & 0.40 & 0.741 \\
tshirt-black-small & 0.650 & 0.60 & 0.863 \\
tshirt-red & 0.656 & 0.80 & 0.749 \\
\midrule
\textit{Average} & 0.574 & 0.625 & 0.704 \\
\midrule
\rowfont{}
\textbf{\textit{\makecell[l]{Overall Average\\(only without *)}}} & 0.583 & 0.675 & 0.695  \\
\textbf{\textit{Overall Average}} & 0.594 & 0.675 & 0.715 \\
\bottomrule
\end{tabu}
\end{table}

\subsection{Evaluation Protocol}
\label{sec:evaluation-protocol}
We evaluated each model on 16 unique items, repeating each evaluation 5 times to account for variability in the initial cloth configuration. {As part of our ablation study, we conducted two sets of experiment runs: the first with four models, and the second with an additional three models.}

Since the initial cloth configuration—specifically, the location where the robot's left hand holds the cloth—strongly influences grasp success, we sought to minimize this variability by evaluating all models within a given run using the same initial grasp point obtained from steps (i) and (ii) defined in Section~\ref{sec:task}. This ensured a consistent gripper position for the first grasp, although the exact cloth configuration was not always replicated because previous grasp trials could slightly alter the cloth state. Nevertheless, we observed that similar configurations were often achieved. To further mitigate cross-model interference, we randomized the order in which models were evaluated within each run.

{
This protocol enables fair comparison of models within the same run; however, comparing models across different runs is less appropriate due to \revfivechanges{the} significant influence of \revfivechanges{the} initial cloth configuration. To provide a reference between runs, we included the best-performing model in both. In total, eight models (seven unique) were evaluated, resulting in 640 grasping trials.
}
\paragraph{Evaluation Items} The evaluation items included 8 objects from the ICRA competition that were provided to us by the organizers (4 towels and 4 t-shirts), as well as 8 additional items that we obtained independently (3 towels and 5 t-shirts)\footnote{Items will be made available to the research community for evaluation purposes.}. All items are depicted in Fig.~\ref{fig:cloth-references}. Out of 16 items, 2 towels and 2 t-shirts from the competition set were also part of the training set, while the remaining 12 items were never observed during training.

\paragraph{Metrics} We measure the cloth coverage metric as used in~\cite{DeGusseme2024BenchmarkingGrasp}, which is defined as the ratio of the stretched-out cloth's surface area to the reference surface area. Surface area is measured \revfivechanges{as} the number of pixels belonging to the cloth\revfivechanges{, with images} captured at a predefined distance from the camera for a fair comparison. The segmentation of the cloth that defines its surface area has been automatically obtained using SAM~\cite{Kirillov2023}, with manual corrections for any incorrect segmentations. We calculated the coverage metric from \revfivechanges{the} two \revfivechanges{stereo-camera} images and report \revfivechanges{their} mean as the final coverage metric. We also measured the number of successful grasps.% and the number of re-tries at different initial positions $Q_{cloths}$.

\subsection{Results}

\begin{figure*}
    \centering
    \includegraphics[width=\linewidth]{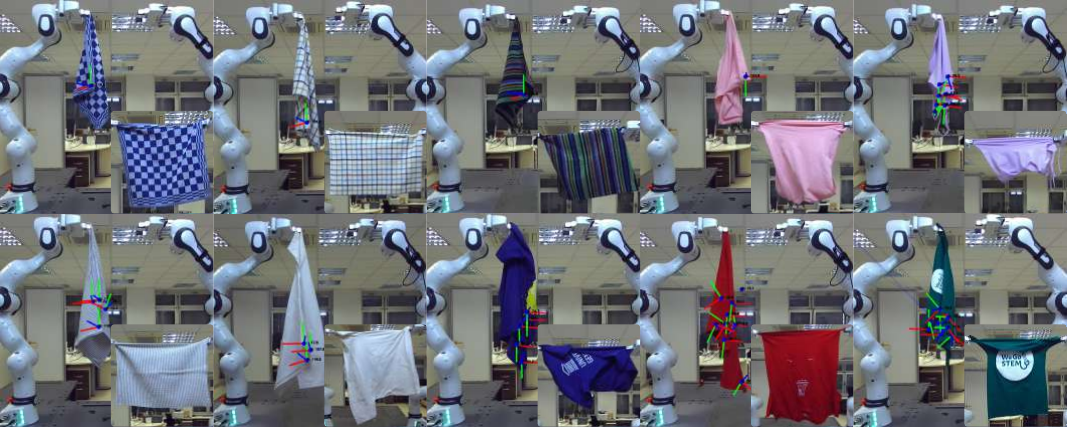}
    \caption{Examples of detected grasp points and gripper poses with the corresponding unfolded item at the \revfivechanges{bottom} left of each image. \revthreechanges{Supplementary video material provides the full execution sequence for all shown examples, including intermediate states between the predicted grasp and the final unfolded outcome.}}
    \label{fig:detection}
\end{figure*}

Detailed results for the \revfivechanges{best-performing} CeDiRNet-6DoF model on individual items from ICRA 2024 and our set are provided in Table~\ref{tab:results_by_item}. { In this table, we report values averaged over two sets of experiment runs for this model (from RUNS A and RUNS B), effectively doubling the number of repetitions and increasing statistical robustness. Some examples of estimated grasp poses and unfolding outcome\revfivechanges{s} are depicted in Fig.~\ref{fig:detection}.}

\begin{figure}
    \centering
    \includegraphics[width=1\linewidth]{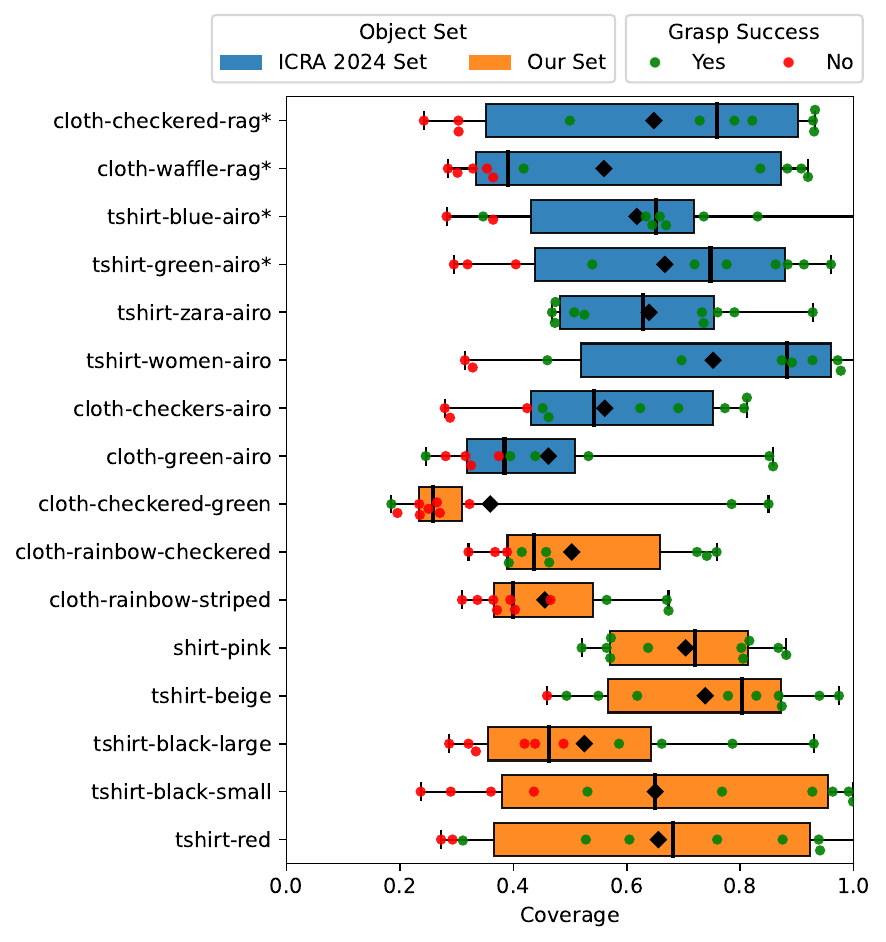}
    \caption{\revfivechanges{Evaluation of} CeDiRNet-6DoF on \revfivechanges{the} ICRA 2024 and our sets. Coverage for ten repeated executions for each item \revfivechanges{is depicted} as green/red circles, with \revfivechanges{the} box-plot showing median (black line), mean (black diamond), min/max (box whiskers) and 75th/25th \revfivechanges{percentiles} (box). Items observed during training are marked with (*).
    }
    \label{fig:variance-by-items}
\end{figure}

On the ICRA 2024 set, the method achieved 61.3\% average coverage and 72.5\% grasp success. Our set of items proved slightly more difficult, with the same model achieving 57.4\% average coverage and 62.5\% grasp success. Note that the coverage metric includes both successful and unsuccessful grasps, with unsuccessful grasps resulting in lower values. For successful grasps, CeDiRNet-6DoF achieved 72.6\% coverage on the ICRA 2024 set and 70.4\% on our set. {Note that half of the items in the ICRA set were included in the training data, whereas none of the items in our set were seen during training. We also provide results considering only novel items which are marked as \textit{(only without *)} in Table~\ref{tab:results_by_item}.}

\begin{table*}
\centering
\caption{Ablation study for CeDiRNet-6DoF with individual components disabled. The best result for each metric is shown in bold {for each set of experiment runs. In both sets, the best results were} mostly achieved with the 6-DoF version using segmentation, random background augmentation\revfivechanges{,} and \revfivechanges{robot-position} cropping.
}
\label{tab:ablation}
\setlength{\tabcolsep}{5.3pt}
\begin{tabu}{m{4pt}|ccccc|ccccccccc}
\toprule
\multirow{2}{*}{} & \multirow{2}{*}{\makecell{\textit{DoF}}} & \multirow{2}{*}{\textit{Seg.}} & \multirow{2}{*}{\textit{\makecell{Random bg.\\augment.}}} & \multirow{2}{*}{\textit{\makecell{Crop.~by\\robot pos.}}} & \multirow{2}{*}{\textit{RGB-D}} & \multicolumn{3}{c}{\textit{Coverage}} & \multicolumn{3}{c}{\textit{\makecell{Grasp Success}}} & \multicolumn{3}{c}{\textit{\makecell{Coverage with Succ. Grasp}}} \\

 & & & & & & \textit{ICRA 2024} & \textit{Our} & \textit{Both} & \textit{ICRA 2024} & \textit{Our} & \textit{Both} & \textit{ICRA 2024} & \textit{Our} & \textit{Both} \\
\toprule

\multirow{4}{*}{\rotatebox{90}{\textbf{RUNS A}}} 
 & 6-DoF & & & &                                     & 0.455 & 0.420 & 0.437 & 0.388 & 0.225 & 0.306 & 0.678 & \textbf{0.735} & 0.709 \\ % v2
 & 6-DoF & \checkmark & & &                          & 0.490 & 0.458 & 0.474 & 0.475 & 0.325 & 0.400 & 0.660 & 0.646 & 0.653 \\ % v2+seg
 & 6-DoF & & \checkmark & \checkmark &               & 0.604 & 0.578 & 0.591 & 0.675 & \textbf{0.625} & 0.650 & 0.698 & 0.706 & 0.702 \\
 & 6-DoF & \checkmark & \checkmark & \checkmark &    & \textbf{0.635} & \textbf{0.583} & \textbf{0.609} & \textbf{0.750} & \textbf{0.625} & \textbf{0.688} & \textbf{0.750} & 0.683 & \textbf{0.717} \\ 
\midrule
\rowfont{}
\multirow{4}{*}{\rotatebox{90}{\textbf{RUNS B}}}
 & 3-DoF &  &  &  & & 0.388 & 0.379 & 0.383 & 0.325 & 0.150 & 0.238 & 0.553 & 0.646 & 0.581 \\ 
 \rowfont{}
 & 6-DoF & \checkmark & \checkmark & \checkmark &    & \textbf{0.594} & \textbf{0.570} & \textbf{0.582} & 0.700 & 0.625 & 0.663 & 0.708 & \textbf{0.726} & \textbf{0.717} \\ 
 \rowfont{}
 & 6-DoF & \checkmark & \checkmark & \checkmark &  \checkmark  & 0.593 & 0.534 & 0.564 & 0.675 & 0.650 & 0.663 & \textbf{0.717} & 0.612 & 0.661 \\ 
 \rowfont{}
 & 6-DoF & \checkmark & \checkmark &  &  \checkmark  & 0.577 & 0.502 & 0.539 & \textbf{0.725} & \textbf{0.675} & \textbf{0.700} &  \textbf{0.717} & 0.569 & 0.643 \\ 

\bottomrule
\end{tabu}
\end{table*}

Among individual items, the method performed best on \textit{tshirt-women-airo} from the ICRA 2024 set, with 75.2\% coverage and 80\% grasp success, and worst on \textit{cloth-checkered-green}, with 35.9\% coverage and 30\% grasp success. \revtwochanges{ For \textit{cloth-checkered-green}, we observed that in many failure cases the grasp location was good (or at least not poor); however, grasp orientation seemed to have been particularly poor.} The strong performance on novel T-shirts is also evident for \textit{tshirt-women-airo}, a substantially different sleeveless shirt not observed during training. The model also performs well on \textit{shirt-pink}, whose appearance differs even further from the training images.

Looking at the variance of individual cases depicted in Fig.~\ref{fig:variance-by-items}, we notice that for \textit{shirt-pink}, \textit{cloth-checkered-green} and \textit{tshirt-beige} the model performs fairly consistently, while variance is much larger for other items. This indicates that the method struggles to grasp the cloth consistently, but successful grasps often achieve coverage above 80\% or even 90\%. We also noticed that unsuccessful grasps often stem from incorrect grasp orientation, while the grasp location appeared suitable. This occurred in 27 of the 52 failed grasps.

\subsection{Computational cost}

Next, we measured the runtime of CeDiRNet-6DoF for the image-processing step in \revfivechanges{Step (}iii), i.e., detection and 6-DoF pose regression. \revthreechanges{Candidate filtering based on path reachability, motion planning, and grasp execution were performed separately on the CPU and are not included here.} We evaluated the ConvNeXt-base model on an NVIDIA A100 40GB and an NVIDIA RTX 2080 Ti 11GB, using 10 measured runs after one warm-up run.

\revthreechanges{Runtime scaled mainly with input resolution (Table~\ref{tab:runtime_resolution}), increasing from 0.27 to 0.81 sec on the A100 and from 0.45 to 1.45 sec on the RTX 2080 Ti when moving from $512\times512$ to $3840\times2160$. At the default resolution of $1024\times1024$, GPU memory usage was between 3 and 3.5 GB, and even the RTX 2080 Ti maintained near-real-time image-processing rates.}

\begin{table}[t]
\centering
\caption{\revthreechanges{Computation\revfivechanges{al} cost \revfivechanges{with respect to} resolution and \revfivechanges{GPU model} (in seconds).}}
\label{tab:runtime_resolution}
\scriptsize
\setlength{\tabcolsep}{3pt}
\begin{tabu}{lcccc}
\toprule
 & \revthreechanges{$512\!\times\!512$} & \revthreechanges{$1024\!\times\!1024$} & \revthreechanges{$1920\!\times\!1080$} & \revthreechanges{$3840\!\times\!2160$}\\
\midrule
\revthreechanges{A100 40GB SXM} & \revthreechanges{$0.27\pm0.01$} & \revthreechanges{$0.34\pm0.01$} & \revthreechanges{$0.41\pm0.01$} & \revthreechanges{$0.81\pm0.02$}\\
\revthreechanges{RTX 2080Ti 11GB} & \revthreechanges{$0.45\pm0.04$} & \revthreechanges{$0.55\pm0.01$} & \revthreechanges{$0.77\pm0.03$} & \revthreechanges{$1.45\pm0.06$}\\
\bottomrule
\end{tabu}
\end{table}

\revthreechanges{Scene complexity had negligible effect on runtime. At $1024\times1024$, a scene with 5 final detections and a higher-complexity scene with 12 final detections required 0.30 vs. 0.30 sec on the A100 and 0.51 vs. 0.49 sec on the RTX 2080 Ti. This is due to the design of CeDiRNet, which regresses all pixel locations at once, making the dominant cost dense image-wide feature extraction and pose regression, largely independent of the final number of detections. Only the final non-maximum suppression and ranking depend on the returned candidate count, and represent a negligible part of the total detection time. Therefore, multiple visible objects in the scene should have little influence on detection speed.}

\subsection{Ablation Study}

Next, we evaluated individual components of CeDiRNet-6DoF to clarify several design choices. {We performed two sets of experiment runs. In the first set, \revfourchanges{denoted \textbf{RUNS A}}, we evaluated a 6-DoF CeDiRNet baseline trained with only the Euler-angle and center-direction-vector losses, and compared it with models using segmentation loss and random-background augmentation with robot-position cropping. In the second set, \revfourchanges{denoted \textbf{RUNS B}}, we evaluated the 3-DoF CeDiRNet from our previous work~\cite{Tabernik2024RAL} as a baseline and evaluated CeDiRNet variants using RGB-D rather than RGB input.} Detailed results are shown in Table~\ref{tab:ablation}.

\begin{figure*}
    \centering
    \includegraphics[width=\linewidth]{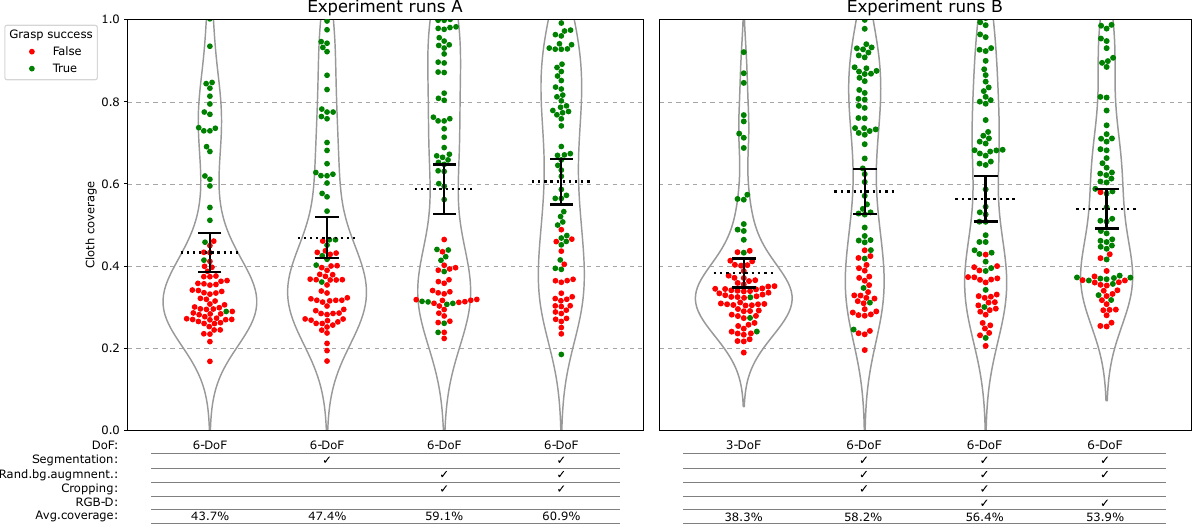}
    \caption{Ablation study for individual CeDiRNet-6DoF components on combined ICRA 2024 and our set, \revfourchanges{with RUNS A/B left/right; dashed/solid black lines show average coverage/95\% CI.}}
    \label{fig:ablation-plot}
\end{figure*}

\paragraph{CeDiRNet-6DoF components}
Focusing on the first set of experiments (\revfourchanges{RUNS A}), the base CeDiRNet model with only the 6-DoF extension, i.e., Euler-angle regression, performed significantly worse than our final CeDiRNet-6DoF. \revfourchanges{For the combined set (``Both'' in Table~\ref{tab:ablation}), the final model improved coverage from 43.7\% to 60.9\% (+17.2 pp.) and grasp success from 30.6\% to 68.8\% (+38.2 pp.).} \revfourchanges{Adding only segmentation improved combined-set coverage from 43.7\% to 47.4\% (+3.7 pp.) and grasp success from 30.6\% to 40.0\% (+9.4 pp.).} This gap can therefore be reduced simply by including segmentation as auxiliary-task supervision.

While segmentation improved performance, the largest gains resulted from random-background augmentation and robot-position cropping. \revfourchanges{Together, these improved combined-set coverage from 43.7\% to 59.1\% (+15.4 pp.) and grasp success from 30.6\% to 65.0\% (+34.4 pp.) over the base model.} %Although we evaluated version with combined random background augmentation and cropping by robot position in a single model, we believe that majority of improvements stem from augmentation with random background. Finally, we also evaluated using RGB-D as input data for CeDiRNet model instead of just RGB. Although this slightly improved grasp success rate by around 5 pp., it then lead to reduced coverage, which is particularly noticeable in coverage for successful grasps.
These performance improvements are also visible in Fig.~\ref{fig:ablation-plot}, which shows coverage for individual samples together with grasp success. Mean coverage is shown {by the dashed black line and its 95\% confidence interval by the solid black lines}. %In particular, we can clearly see poor grasp success rate for base CeDiRNet model, while looking at the RGB-D version we can see several green dots below 50\% coverage, confirming that despite better grasp success rate the model cannot find as optimal points as RGB version.

\paragraph{Depth}
In our second set of runs (\revfourchanges{RUNS B}), we also evaluated models with an additional depth input channel (RGB-D). They are shown in the bottom two rows of Table~\ref{tab:ablation} and the right-most columns of Fig.~\ref{fig:ablation-plot}. \revfourchanges{For matched variants with segmentation, random-background augmentation, and robot-aware cropping, adding depth changed combined-set coverage from 58.2\% to 56.4\% ($-1.8$ pp.) while grasp success remained 66.3\%. The unmatched RGB-D variant achieved 53.9\% coverage and 70.0\% grasp success.} The difference between the matched depth ($\text{56.4}\pm\text{5.5\%}_{\text{95\% CI}}$) and non-depth models ($\text{58.2}\pm\text{5.5\%}_\text{95\% CI}$) was not statistically significant under a $t$-test at $p = 0.05$. We therefore cannot conclude that depth input degrades performance.

Note that the RGB-only approach still relies on depth at inference. The network predicts a 5-DoF grasp pose (two image coordinates and three 3D orientation angles), while the RGB-D camera provides the third positional coordinate.

\begin{figure}
    \centering
    \includegraphics[width=\linewidth]{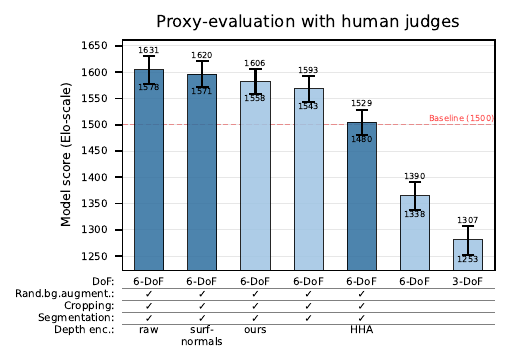}
    \caption{\revtwochanges{Proxy evaluation of depth encodings with human judges. We report ELO-like scores obtained by fitting a Bayesian Bradley--Terry model (Bernoulli likelihood; ties treated as 0.5), together with 95\% credible intervals. Models also evaluated on real robots are shown in lighter blue.}}
    \label{fig:proxy_eval_judges}
\end{figure}

\paragraph{\revtwochanges{Depth encodings}} \revtwochanges{We further compared multiple depth encodings using human judges in the proxy evaluation described below. We evaluated three common depth encodings: a) HHA\footnote{Horizontal disparity, Height above ground, Angle with gravity.}~\cite{Gupta2014}, b) surface normals, and c) raw distances. Results are reported in Fig.~\ref{fig:proxy_eval_judges}. HHA performs worse than the no-depth and other encodings, with its 95\% credible interval clearly below the best-performing models. In contrast, raw distances and surface normals achieve scores comparable to our initial encoding and to the model without depth. While their mean scores are slightly higher, the 95\% credible intervals overlap, so we cannot draw a strong conclusion that these encodings improve performance.}

We performed this evaluation as a model-vs-model benchmark, which serves as a proxy \revfivechanges{f}o\revfivechanges{r} real-robot evaluation. We used five judges to indicate which model's predicted grasp would lead to better unfolding coverage for the same image (considering \revfivechanges{each} grasp's location and 3D orientation), and then aggregated pairwise preferences into ELO-like scores using a Bayesian Bradley--Terry model~\cite{Caron2012}. We treated ties as 0.5 in a Bernoulli formulation and report 95\% credible intervals to assess uncertainty. This benchmark was performed on 100 independently captured images from our robot setup. To relate the proxy evaluation to real-robot results, we included four reference models that were also evaluated on the robot in the ablation study; the proxy ranking matched their real-robot ranking.

\begin{table}
    \centering
    \setlength{\tabcolsep}{4pt}
    \begin{tabular}{llcc}
        \toprule
         \textit{\textbf{Rank}} & \textit{\textbf{Team Name/Model}} & \textit{\makecell{Coverage\\ICRA 2024 setup}} & \textit{\makecell{Coverage\\Our setup}} \\
        \midrule
         1st & AIR-JNU  & \textbf{0.60} & - \\
         2nd & \textbf{CeDiRNet-6DoF (our)}  & 0.57 & $\text{0.61}\pm\text{0.05}_{\text{95\% CI}}$ \\
         3rd & EWHA-glab  & 0.55 & - \\
         4th & SCUT-ROBOT  & 0.53 & - \\
         5th & Team Greater Bay  & 0.53 & - \\
         6th & Samsung Research China  & 0.48 & - \\
         7th & Shibata Lab  & 0.46 & - \\
         8th & AI\&ROBOT LAB  & 0.45 & - \\
         9th & UOS-Robotics  & 0.39 & - \\
         10th & AIS Shinshu  & 0.37 & - \\
         11th & 3C1S  & 0.35 & - \\
        \bottomrule
    \end{tabular}
    \caption{Results from \revfivechanges{the} ICRA 2024 cloth competition~\cite{DeGusseme2024BenchmarkingGrasp} \revfivechanges{comparing} our CeDiRNet-6DoF entry \revfivechanges{with} other participants and \revfivechanges{with} our setup on \revfivechanges{the} ICRA 2024 item set.}
    \label{tab:competition_results}
\end{table}

\paragraph{Baseline CeDiRNet-3DoF}
We additionally evaluated our previous 3-DoF version of CeDiRNet~\cite{Tabernik2024RAL} as a baseline in \revfourchanges{RUNS B}. \revtwochanges{We applied the CeDiRNet-3DoF baseline to our 6-DoF setup using only the minimal execution constraints required to obtain a valid 6-DoF robot pose. This enables a comparison with our 6-DoF variant that stays close to the original 3-DoF formulation and provides evidence for the benefit of moving from 1D to 3D orientation regression.} Since CeDiRNet-3DoF regresses only one rotational degree of freedom, we regress only the yaw of the grasp poses in \revfivechanges{the} world coordinate frame, i.e., the rotation around the world Z-axis (vertical axis). The remaining two rotations (pitch and roll) were constrained such that the gripper always approached the object in an orientation that was parallel to the table surface. Depth information from the RGB-D camera was used to position the grasp pose in 3D world coordinates, similar to the 6-DoF version. \revtwochanges{Note that regressing yaw-only instead of pitch- or roll-only \revfivechanges{preserves} unfolded-coverage \revfivechanges{capability when} grasping hanging cloth, as w\revfivechanges{e} manually verified on \revfivechanges{the} robot with \revfivechanges{human-annotated} grasp poses \revfivechanges{under} the same \revfivechanges{constraints}, whereas regressing pitch- or roll-only would lead to \revfivechanges{a} loss of coverage \revfivechanges{capability}.}

Results for this baseline are presented in the first column of \revfourchanges{RUNS B} on the right in Fig.~\ref{fig:ablation-plot}. The performance was significantly worse than that of our proposed CeDiRNet-6DoF model under a $t$-test at the 0.05 significance level ($\text{38.3}\pm\text{3.5\%}_\text{95\% CI}$ vs. $\text{58.2}\pm\text{5.5\%}_\text{95\% CI}$), demonstrating that regressing all three Euler angles, together with the other proposed improvements, substantially enhances performance.

\subsection{Comparing with State-of-the-Art}

\revthreechanges{Finally, we compare our model with related work. We first provide \revfivechanges{a} comparison \revfivechanges{with} other participants from the ICRA 2024 cloth competition~\cite{DeGusseme2024BenchmarkingGrasp} and \revfivechanges{then} provide \revfivechanges{a human-judge} comparison \revfivechanges{with} two related models.}

\paragraph{ICRA 2024 cloth competition} \revthreechanges{We submitted our best CeDiRNet-6DoF model without depth}, i.e., a version with segmentation, random background augmentation, \revfivechanges{robot-aware} cropping\revfivechanges{,} and RGB input. The main difference \revfivechanges{from} our setup was primarily in \revfivechanges{the} different robot arms and in \revfivechanges{the} number of repetitions, {only \revfivechanges{two} for the competition and \revfivechanges{ten} in our case, as we report results from \revfivechanges{the} combined \revfourchanges{RUNS A and RUNS B}.}

{Detailed results of the competition are reported in~\cite{DeGusseme2024BenchmarkingGrasp}, while we summarize the results} in Table~\ref{tab:competition_results}. Our model achieved coverage of 57\%\revfivechanges{,} which resulted in second place. CeDiRNet-6DoF \revfivechanges{outperformed nine} other methods, except the method by AIR-JNU which achieved coverage of 60\%. \revfivechanges{In} our setup, CeDiRNet-6DoF achieved {$\text{61}\%\pm\text{5}_{\text{95\% CI}}$} coverage using the identical set of items, which is 4 pp. better than \revfivechanges{in the} ICRA 2024 competition despite using the exact same model. However, ICRA 2024 results also include two samples that were disqualified due to technical issues, which represent\revfivechanges{ed} 12.5\% of all examples due to \revfivechanges{the} small number of repetitions. Despite this disadvantage, the model achieved second place and lagged behind the \revfivechanges{first-ranked method} only by 3 pp. \revtwochanges{Without the two samples, the method would have achieved 59.01\% coverage, \revfourchanges{0.99 pp. below the first-ranked result of 60\%}, thus indicating that CeDiRNet-6DoF achieves results comparable to state-of-the-art. }

\begin{figure}[t]
    \centering
    \includegraphics[width=\linewidth]{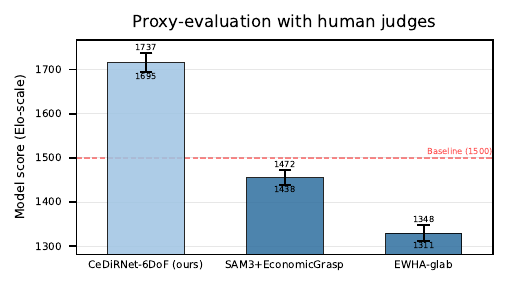}
    \caption{\revthreechanges{Proxy evaluation against two related-work baselines (in dark blue) with human judges. Reporting ELO-like score from fitted hierarchical Bayesian Bradley--Terry model and 95\% credible intervals.}}
    \label{fig:proxy_eval_relatedwork}
\end{figure}

\revthreechanges{\paragraph{Human judges evaluation} To complement the competition ranking, we also extended the proxy evaluation to related-work baselines using the same protocol as in the depth ablation study. Out of the cloth-specific and rigid 6-DoF grasp-pose methods considered in Related Work, we evaluated the two with directly reproducible public implementations for ICRA-style benchmarking: the rigid-object 6-DoF detector EconomicGrasp~\cite{wu2025economic} and the EWHA-glab method from the ICRA 2024 cloth competition. For EconomicGrasp, this required an explicit adaptation to our cloth setting by segmenting the hanging cloth with SegmentAnything3, restricting grasp regression to the segmented cloth region, and training a best-performing adapted model on the ICRA 2024 competition training set. EWHA-glab was evaluated using its released public code \footnote{\url{https://github.com/minseo10/cloth_unfolding}} and did not require any training.}

\revthreechanges{Using the same 100 evaluation samples and the same five judges as in the previous proxy benchmark, our best CeDiRNet-6DoF variant with our initial depth encoding, segmentation, and robot-aware cropping achieved an ELO-like score of 1715.0, compared with 1455.2 for SAM3+EconomicGrasp and 1329.8 for EWHA-glab. \revfourchanges{Figure~\ref{fig:proxy_eval_relatedwork} shows this comparison.}}

\section{\revfourchanges{Discussion}}

\revfourchanges{Overall, the results support CeDiRNet-6DoF and its main design choices for robot cloth unfolding, validated by gains of 17.2 pp. for coverage and 38.2 pp. for grasp success on the combined evaluation set. It also placed second in the ICRA 2024 cloth competition, outperforming nine participants, and ranked highest in our three-method proxy comparison with reproducible related-work baselines. Its advantage over EconomicGrasp, which learns grasp poses from point clouds by densely predicting objectness, graspness, angle, and depth, also supports the center-direction vectors and direct dense grasp-prediction formulation used by CeDiRNet-6DoF.}

\paragraph{\revfourchanges{Generalization}}
\revfourchanges{Results on items excluded from training support generalization to the evaluated unseen towels and T-shirts, while other garment categories remain to be evaluated. At the item level, performance was higher on the novel T-shirts than on the novel cloths. This may reflect differences in appearance: the novel cloths had substantially different textures, whereas variation in the novel T-shirts was largely limited to central logos. Likewise, although checkered patterns occurred in training, they may not have transferred to orientation prediction for the novel \textit{cloth-checkered-green}.}

\paragraph{\revfourchanges{Auxiliary segmentation}}
\revfourchanges{Although the segmentation head output is not used at inference time, its loss provides auxiliary supervision during training. The measured improvement from using segmentation in training may arise because this supervision encourages shared features useful for both segmentation and grasp-pose estimation, thereby having a regularization-like effect~\cite{Caruana1997,Szegedy2014}. We emphasize that this is a plausible explanation rather than a confirmed mechanism.}

\paragraph{\revfourchanges{Depth input}}
\revfourchanges{Depth might be expected to improve grasp-pose estimation where 3D geometry matters, especially given the improvement in our prior 3-DoF evaluation~\cite{Tabernik2024RAL}. Its lack of benefit here may reflect the smaller fine-tuning set (approximately 500 images versus more than 5000 real images for the prior 3-DoF model), which could promote overfitting in the depth channel of the ConvNeXt backbone with more than 88 million parameters. Depth augmentation and randomization, such as dropout or noise injection, might help but were not evaluated. Alternatively, RGB appearance may already provide useful grasp cues, such as texture near cloth corners, and ImageNet pre-training provides strong RGB priors; learning general depth-specific features may require more data or \revfivechanges{benefit from} specialized architectures~\revfiveremovedcite{Wang2021g}\cite{Yin2024}.}

\section{Conclusion}

In this work, we introduced CeDiRNet-6DoF—a novel deep-learning framework for grasp pose estimation \revfivechanges{with a} full 6-DoF representation for robotic cloth manipulation. Our approach integrates dense regression of 2D grasp locations and 3D orientation angles with segmentation and robust data augmentation techniques, while camera depth provides the third positional coordinate, enabling effective grasp pose detection and execution in dual-arm cloth unfolding tasks. We embed this approach within a complete pipeline for dual-arm robotic grasping, leveraging analytical inverse kinematics and an efficient RRT-Connect planner within a regrasping-in-the-air strategy.

Our experimental setup, \revfivechanges{which} closely replicated the ICRA 2024 competition, demonstrated \revfivechanges{through} an extensive evaluation on the original eight ICRA 2024 items as well as on our {previously unobserved} eight items that CeDiRNet-6DoF achieves \revtwochanges{results comparable to state-of-the-art} for this task. %The detailed ablation study provides further evidence for our design choices, underscoring the benefits of incorporating segmentation loss, random background augmentation, and image cropping based on robot positions—in total contributing significantly to the overall performance.
The detailed ablation study provides further evidence for our design choices. {We explored the influence of segmentation loss on both RGB and RGB-D models\revfivechanges{,} demonstrating \revfivechanges{a} slight benefit from \revfivechanges{adjusting} features for segmentation. Comparison \revfivechanges{with the} prior 3-DoF model demonstrated the \revfivechanges{benefit} of \revfivechanges{a three-dimensional} orientation loss \revfivechanges{over a single-dimensional} one.} Furthermore, random background augmentation and image cropping based on robot positions are shown to significantly contribute to the overall performance. Although the task's inherent difficulty is reflected in a high variance of results, the successful cases indicate that our system generalizes well to diverse cloth types and configurations. \revtwochanges{However, our experimental validation currently focuses only on towels and T-shirts. Because the grasp-pose predictor is fully learning-based and does not rely on garment-specific heuristics, the same method can in principle be extended to garments with substantially different geometries (e.g., trousers or long-sleeved shirts) without changing the formulation itself. However, this would require additional training data and dedicated evaluation, and we have not validated such generalization in this work.}

Notably, our model secured second place in the ICRA 2024 cloth competition by achieving 57\% average coverage and outperforming nine competing methods from the challenge. Although the winning method attained 60\% coverage, our subsequent evaluation on the identical set of items using the exact same method without any technical issues yielded 61\% average coverage. \revfivechanges{These results show} that our approach \revfivechanges{performs comparably} to \revfivechanges{the best competition entry}. %Notably, our model secured second place in the ICRA 2024 cloth competition by achieving 57\% average coverage and outperforming nine state-of-the-art methods. While in the competition CeDiRNet-6DoF lagged slightly behind the first place with 60\% average coverage, we achieved even higher coverage at 61\% with our setup using the exact same set of items and the same method, indicating that CeDiRNet-6DoF is at least comparable if not even outperforming all other competing methods.

Overall, our findings confirm that the proposed framework not only advances the state-of-the-art in cloth manipulation but also offers a promising foundation for future research. \revthreechanges{At the same time, the current formulation is multi-candidate at the grasp-selection level, but deterministic at the per-candidate level: it does not estimate calibrated uncertainty or multiple pose hypotheses for the same grasp location. This design is consistent with the benchmark, which provides a single annotated 6-DoF grasp per image and requires execution of a single final grasp, but richer uncertainty-aware or multi-hypothesis grasp prediction could be beneficial in ambiguous cloth configurations.} Future work could focus on improving grasp-pose orientation, which has often proven to be the main issue, while also expanding the method to handle even more diverse and challenging items (\revfivechanges{long-sleeved shirts}, trousers, etc.). \revthreechanges{More geometry-aware rotation representations, such as quaternion-based parameterizations or other rotation-aware losses, also remain worthwhile future directions to help improve grasp pose orientation accuracy.} {  The currently used benchmarking protocol also offers only limited environmental variability and tends to reflect the performance of methods under more controlled settings such as in factories. To advance benchmarking in 6‑DoF cloth grasping, more challenging environment\revfivechanges{al variation} should be introduced in the future, for example, varying backgrounds, different lighting conditions, and a wider range of camera viewpoints.}

\section{Acknowledgment}

This work was supported by the ARIS research project J2-4457 (RTFM) and programme groups P2-0214 and P2-0076. It also received funding from the European Union's Horizon Europe Framework Programme under Grant Agreement No. 101159522 (ROMANDIC).

%\vspace{-1.5\baselineskip}
\bibliographystyle{IEEEtran}
%\bibliography{bibtex-domen}
\bibliography{bibtex-domen_no_urls}
%\bibliography{literatura-robotika}

% Compact only the required post-acceptance author biographies. IEEEtran's
% default four-baseline gap before every biography leaves excessive whitespace.
\makeatletter
\def\@IEEEBIOskipN{0pt}
\makeatother

\begin{IEEEbiography}[{\includegraphics[width=1in,height=1.25in,clip,keepaspectratio]{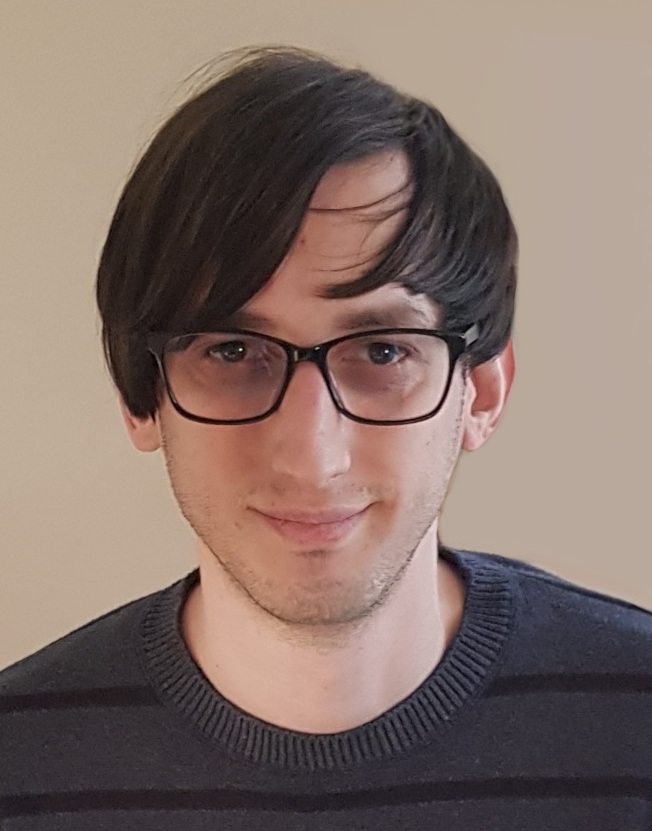}}]{Domen Tabernik} is a Research Associate with the Visual Cognitive Systems Laboratory, Faculty of Computer and Information Science, University of Ljubljana, Slovenia. His research focuses on deep learning methods for computer vision, with current work spanning visual anomaly detection, industrial surface inspection, point-supervised object localization, and vision-based robotic manipulation. His research interests include computer vision, deep learning, industrial inspection, and robotic perception.
\end{IEEEbiography}

\begin{IEEEbiography}[{\includegraphics[width=1in,height=1.25in,clip,keepaspectratio]{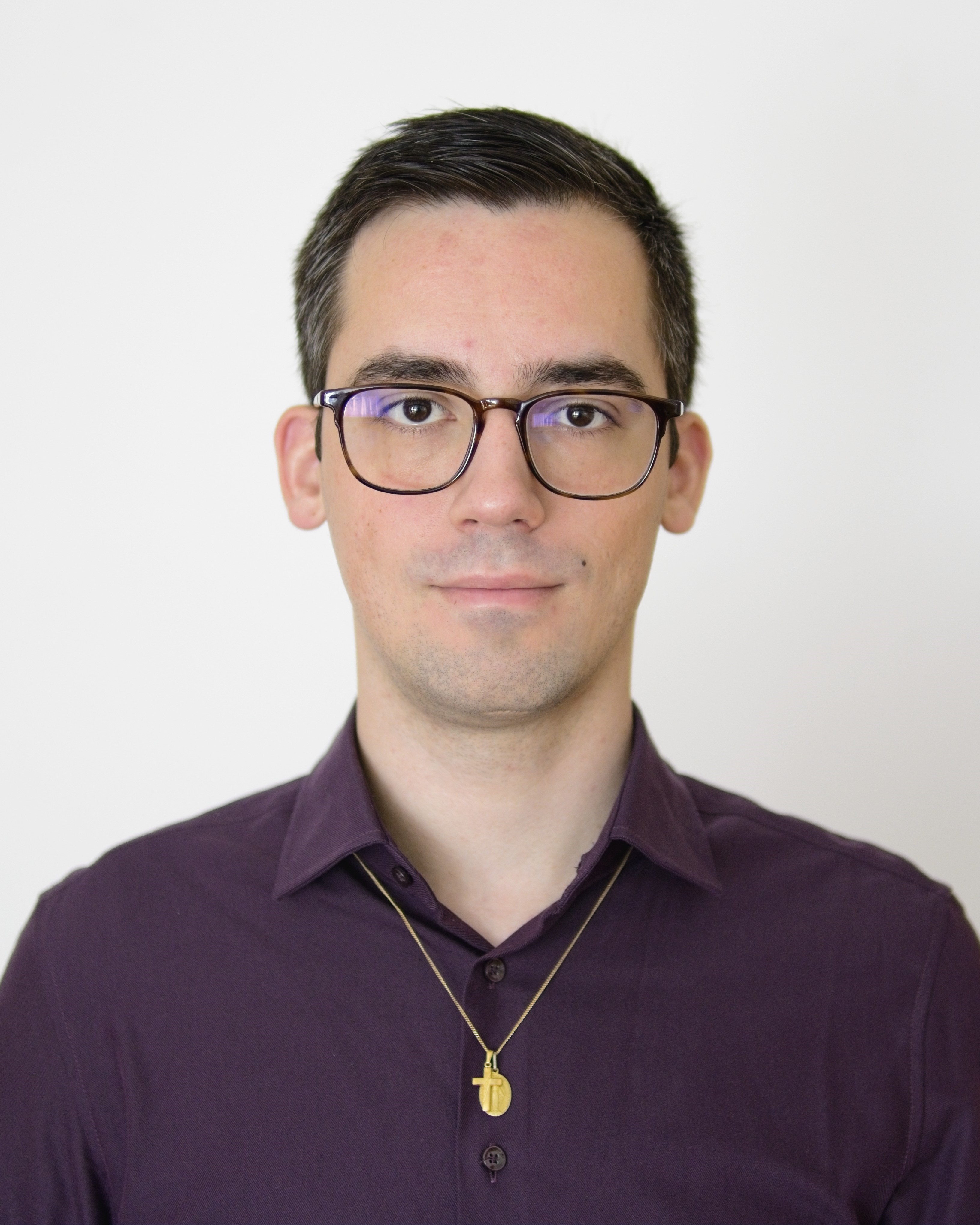}}]{Peter Nimac}
is a Ph.D. student with the Department of Automatics, Biocybernetics and Robotics, Jo\v{z}ef Stefan Institute, Ljubljana, Slovenia, and is pursuing his doctoral studies at the Jo\v{z}ef Stefan International Postgraduate School. His current work focuses on robotic cloth manipulation. His research interests include robotics, radar sensing, object tracking, and human--robot collaboration.
\end{IEEEbiography}

\begin{IEEEbiography}[{\includegraphics[width=1in,height=1.25in,clip,keepaspectratio]{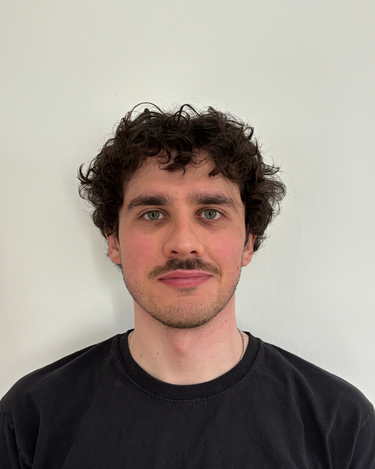}}]{Jan Jeri\'cevi\'c}
is a Ph.D. student with the Department of Automatics, Biocybernetics and Robotics, Jo\v{z}ef Stefan Institute, Ljubljana, Slovenia, and is pursuing his doctoral studies at the Jo\v{z}ef Stefan International Postgraduate School. His current work focuses on robotic cloth manipulation and capability-guided task execution. His research interests include deformable object manipulation, manipulability and workspace analysis, and composable manipulation primitives.
\end{IEEEbiography}

\begin{IEEEbiography}[{\includegraphics[width=1in,height=1.25in,clip,keepaspectratio]{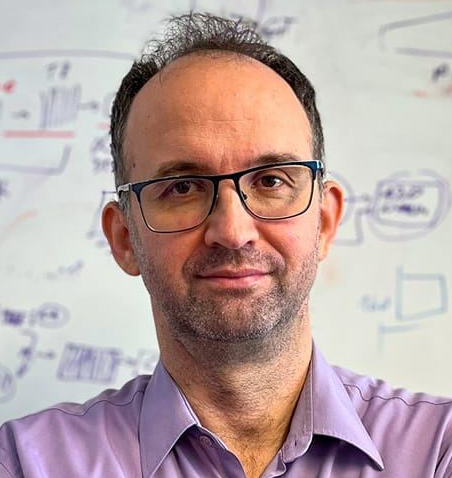}}]{Danijel Sko\v{c}aj}
is a Full Professor and Head of the Visual Cognitive Systems Laboratory at the University of Ljubljana, Faculty of Computer and Information Science. His main research interests include computer vision, deep learning, and cognitive robotics. He has led numerous research projects in these areas and has actively contributed to the transfer of research results into practical applications. He has served as President of the IEEE Slovenia Computer Society and President of the Slovenian Pattern Recognition Society.
\end{IEEEbiography}

\begin{IEEEbiography}[{\includegraphics[width=1in,height=1.25in,clip,keepaspectratio]{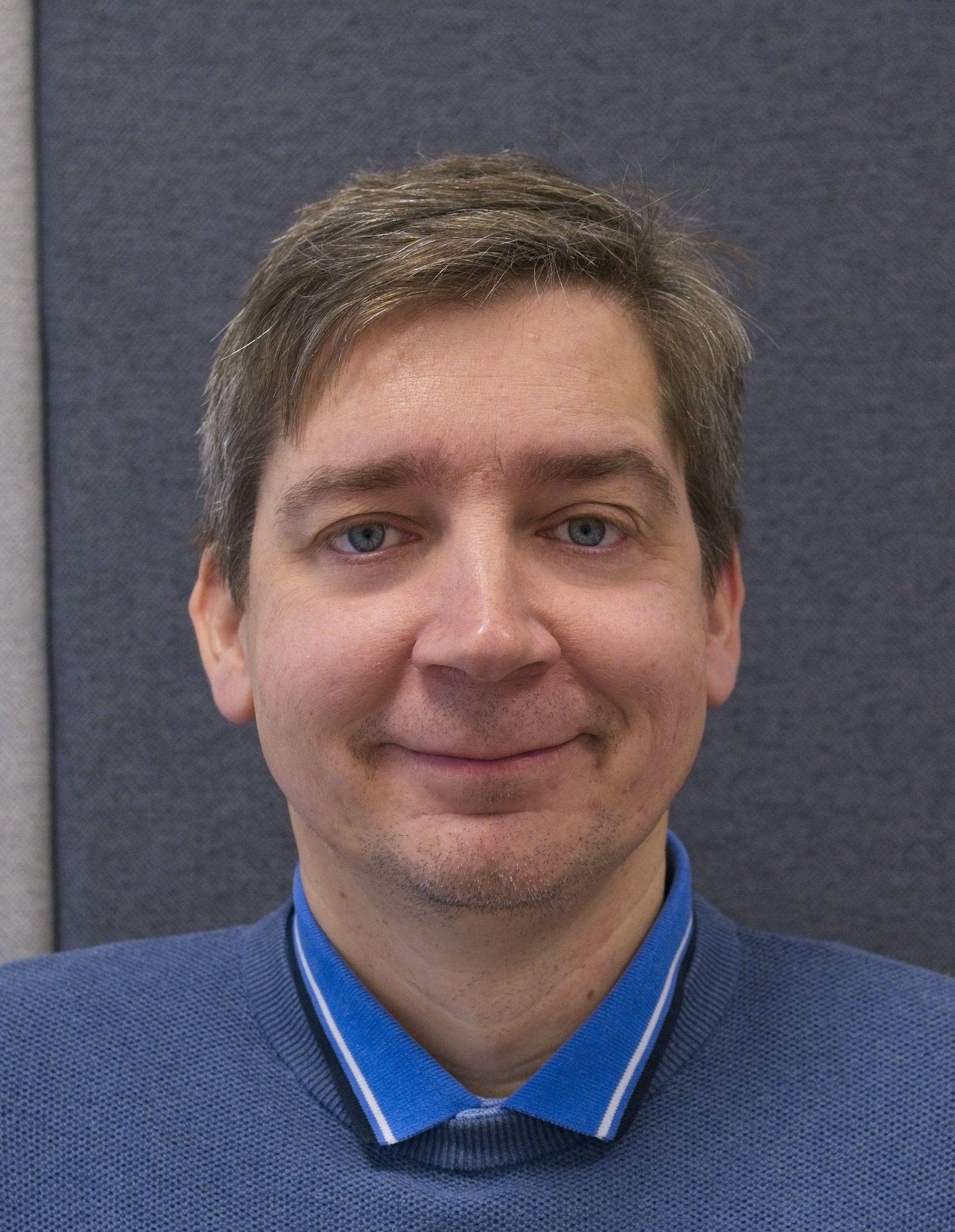}}]{Andrej Gams}
is a Senior Research Associate with the Department of Automatics, Biocybernetics and Robotics, Jo\v{z}ef Stefan Institute, Ljubljana, Slovenia, where he heads the Humanoid and Cognitive Robotics Laboratory. He is also an Associate Professor with the Jo\v{z}ef Stefan International Postgraduate School. His current work and research interests focus on robot learning and robotic manipulation of deformable objects, with applications in humanoid and industrial robotics.
\end{IEEEbiography}

\end{document}